\documentclass[11pt,a4paper]{article}
\usepackage[hyperref]{emnlp2020}
\usepackage{times}
\usepackage{latexsym}
\usepackage{paralist}

\usepackage{microtype}

\usepackage{color}
\usepackage{multirow}
\usepackage{amsmath}
\usepackage{subfig}
\usepackage{enumitem}
\usepackage{amsfonts}
\usepackage{makecell}
\usepackage{arydshln}
\usepackage{multicol}
\usepackage{graphicx}
\usepackage{graphics}
\usepackage{balance}
\usepackage{subfig}
\usepackage{framed}

\aclfinalcopy 

\definecolor{forestgreen}{rgb}{0.13, 0.55, 0.13}

\definecolor{zptu}{RGB}{18, 141, 21}

\title{On the Sub-layer Functionalities of Transformer Decoder}

\author{
Yilin Yang\thanks{~~Work done when interning at Tencent AI Lab.} \\ \normalsize Oregon State University \\  \small {\sf yilinyang721@gmail.com}   \And
Longyue Wang \\ \normalsize Tencent AI Lab \\ \small {\sf vinnylywang@tencent.com} \AND
Shuming Shi \\ \normalsize Tencent AI Lab \\ \small {\sf shumingshi@tencent.com}   \And
Prasad Tadepalli \\ \normalsize Oregon State University \\ \small {\sf tadepall@oregonstate.edu}   \And
Stefan Lee \\ \normalsize Oregon State University \\ \small {\sf leestef@oregonstate.edu}   \And
Zhaopeng Tu \\ \normalsize Tencent AI Lab \\ \small {\sf zptu@tencent.com}
}

\date{}

\begin{document}
\maketitle
\begin{abstract}
There have been significant efforts to interpret the encoder of Transformer-based encoder-decoder architectures for neural machine translation (NMT); meanwhile, the decoder remains largely unexamined despite its critical role. During translation, the decoder must predict output tokens by considering both the source-language text from the encoder and the target-language prefix produced in previous steps. In this work, we study how Transformer-based decoders leverage information from the source and target languages -- developing a universal probe task to assess how information is propagated through each module of each decoder layer. We perform extensive experiments on three major translation datasets (WMT En-De, En-Fr, and En-Zh). Our analysis provides insight on when and where decoders leverage different sources. Based on these insights, we demonstrate that the residual feed-forward module in each Transformer decoder layer can be dropped with minimal loss of performance -- a significant reduction in computation and number of parameters, and consequently a significant boost to both training and inference speed.

\end{abstract}

\section{Introduction}
Transformer models have advanced the state-of-the-art on a variety of natural language processing (NLP) tasks, including machine translation~\cite{Vaswani:2017:NIPS}, natural language inference~\cite{Shen:2018:AAAI}, semantic role labeling~\cite{strubell2018linguistically}, and language representation~\cite{devlin2018bert}.
However, so far not much is known about the internal properties and functionalities it learns to achieve its superior performance, which poses significant challenges for human understanding of the model and potentially designing better architectures.

Recent efforts on interpreting Transformer models mainly focus on assessing the encoder representations~\cite{raganato2018analysis,yang2019assessing,tang:2019:emnlp} or interpreting the multi-head self-attentions~\cite{li2018multi,voita2019analyzing,Michel:2019:NeurIPS}. At the same time, there have been few attempts to interpret the decoder side, which we believe is also of great interest, and should be taken into account while explaining the encoder-decoder networks.
The reasons are threefold:
(a) the decoder takes both source and target as input, and implicitly performs the functionalities of both alignment and language modeling, which are at the core of machine translation;
(b) the encoder and decoder are tightly coupled in that the output of the encoder is fed to the decoder and the training signals for the encoder are back-propagated from the decoder; and (c)
recent studies have shown that the boundary between the encoder and decoder is blurry, since some of the encoder functionalities can be substituted by the decoder cross-attention modules~\cite{Tang:2019:RANLP}.

In this study, we interpret the Transformer decoder by investigating when and where the decoder utilizes source or target information across its stacking modules and layers.
Without loss of generality, we focus on the {\it representation evolution}\footnote{By ``evolution'', we denote the progressive trend from the first layer till the last.} within a Transformer decoder.
To this end, we introduce a novel sub-layer\footnote{Throughout this paper, we use the terms ``sub-layer'' and ``module'' interchangeably.} split with respect to their functionalities:
{\em Target Exploitation Module} (TEM) for exploiting the representation from translation history, {\em Source Exploitation Module} (SEM) for exploiting the source-side representation, and {\em Information Fusion Module} (IFM) to combine representations from the other two (\S\ref{sec:partition}).

Further, we design a universal probing scheme to quantify the amount of specific information embedded in network representations.
By probing both source and target information from decoder sub-layers, and by analyzing the alignment error rate (AER) and source coverage rate, we arrive at the following findings:
\begin{itemize}
    \item SEM guides the representation evolution within NMT decoder (\S\ref{sec:information-quantify}).
    \item Higher-layer SEMs accomplish the functionality of word alignment, while lower-layer ones construct the necessary contexts (\S\ref{sec:source-exploitation}).
    \item TEMs are critical to helping SEM build word alignments, while their stacking order is not essential (\S\ref{sec:source-exploitation}).
\end{itemize}
Last but not least, we conduct a fine-grained analysis on the information fusion process within IFM.
Our key contributions in this work are:
\begin{itemize}[leftmargin=0.85em]
	\item[1.] We introduce a novel sub-layer split of Transformer decoder with respect to their functionalities.
    \item [2.] We introduce a universal probing scheme from which we derive aforementioned conclusions about the Transformer decoder.
    \item [3.] Surprisingly, we find that the de-facto usage of residual FeedForward operations are not efficient, and could be removed in totality with minimal loss of performance, while significantly boosting the training and inference speeds.
\end{itemize}

\section{Preliminaries}

\subsection{Transformer Decoder}

NMT models employ an encoder-decoder architecture to accomplish the translation process in an end-to-end manner. The encoder transforms the source sentence into a sequence of representations, and the decoder generates target words by dynamically attending to the source representations. Typically, this framework can be implemented with a recurrent neural network (RNN)~\cite{Bahdanau:2015:ICLR}, a convolutional neural network (CNN)~\cite{gehring2017convolutional}, or a Transformer~\cite{Vaswani:2017:NIPS}.
We focus on the Transformer architecture, since it has become the state-of-the-art model on machine translation tasks, as well as various text understanding~\cite{devlin2018bert} and generation~\cite{radford2019language} tasks.

Specifically, the decoder is composed of a stack of $N$ identical layers, each of which has three sub-layers, as illustrated in Figure~\ref{fig:architecture}. A residual connection~\cite{he2016deep} is employed around each of the three sub-layers, followed by layer normalization~\cite{ba2016layer} (``Add \& Norm'').
The first sub-layer is a self-attention module that performs self-attention over the previous decoder layer:
\begin{eqnarray}
    {\bf C}_d^n &=& \textsc{Ln}\big(\textsc{Att}({\bf Q}_d^{n}, {\bf K}_d^{n}, {\bf V}_d^{n}) + {\bf L}_d^{n-1} \big) \nonumber
\end{eqnarray}
where $\textsc{Att}(\cdot)$ and $\textsc{Ln}(\cdot)$ denote the self-attention mechanism and layer normalization. ${\bf Q}_d^n$, ${\bf K}_d^{n}$, and ${\bf V}_d^{n}$ are query, key and value vectors that are transformed from the ({\em n-1})-th layer representation ${\bf L}_d^{n-1}$.
The second sub-layer performs attention over the output of the encoder representation:
\begin{eqnarray}
    {\bf D}_d^n &=& \textsc{Ln}\big(\textsc{Att}({\bf C}_d^{n}, {\bf K}_e^N, {\bf V}_e^N) + {\bf C}_d^{n} \big) \nonumber
\end{eqnarray}
where ${\bf K}_e^{N}$ and ${\bf V}_e^{N}$ are transformed from the top encoder representation ${\bf L}^{N}_{e}$. The final sub-layer is a position-wise fully connected feed-forward network with ReLU activations:
\begin{eqnarray}
    {\bf L}_d^n &=& \textsc{Ln}\big(\textsc{Ffn}({\bf D}_d^n) + {\bf D}_d^{n} \big) \nonumber
\end{eqnarray}
The top decoder representation ${\bf L}_d^N$ is then used to generate the final prediction.

\subsection{Sub-Layer Partition}
\label{sec:partition}

\begin{figure}[t]
	\centering
	\includegraphics[width=0.48\textwidth]{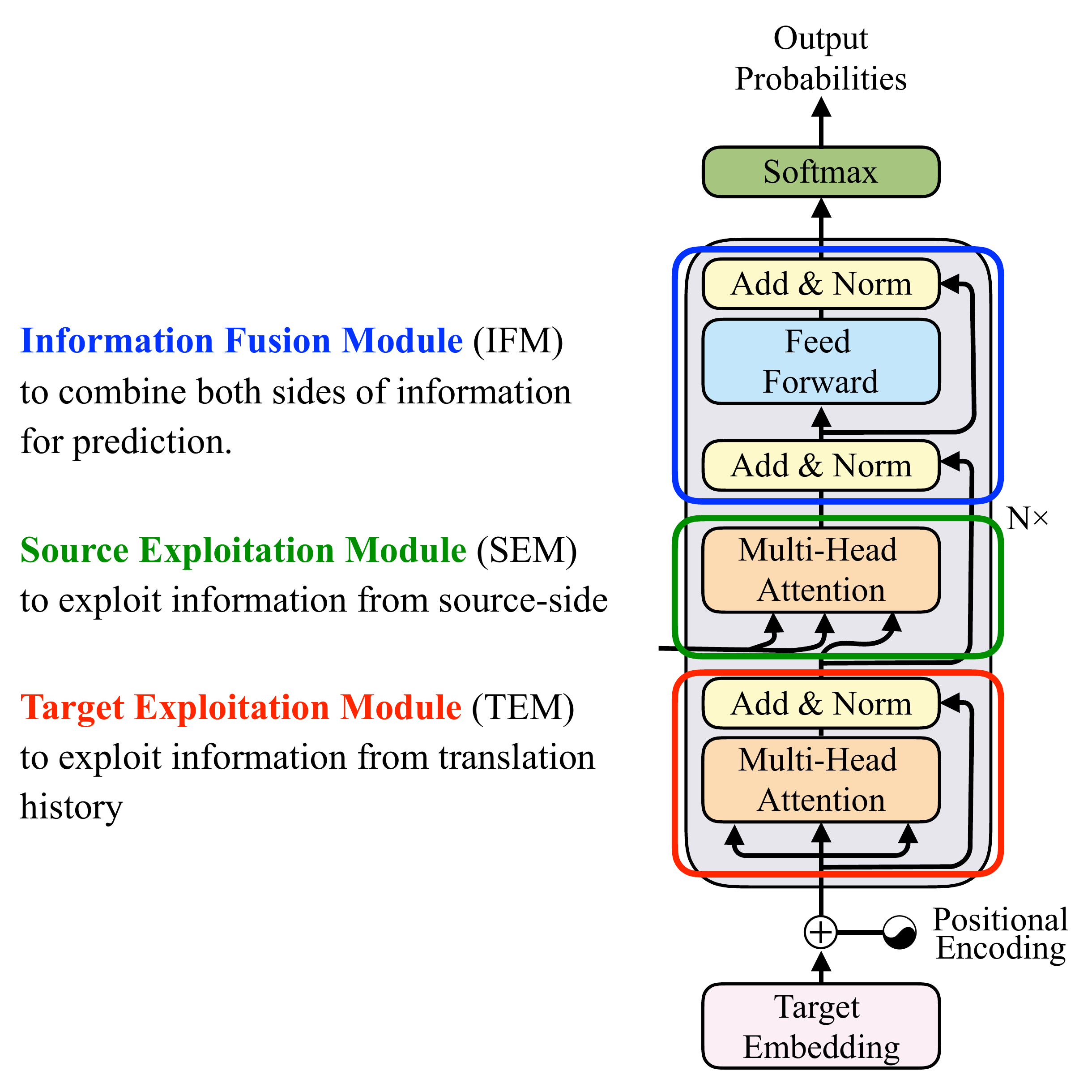}
	\caption{A sub-layer splitting of Transformer decoder with respect to their functionalities.}
	\label{fig:architecture}
\end{figure}

In this work, we aim to reveal how a Transformer decoder accomplishes the translation process utilizing both source and target inputs.
To this end, we split each decoder layer into three modules with respect to their different functionalities over the source or target inputs, as illustrated in Figure~\ref{fig:architecture}:
\begin{itemize}[leftmargin=0.85em]
    \item {\em Target Exploitation Module} (TEM) consists of the self-attention operation and a residual connection, which exploits the target-side translation history from previous layer representations. 
    \item {\em Source Exploitation Module} (SEM) consists only of the encoder attention, which dynamically selects relevant source-side information for generation.
    \item {\em Information Fusion Module} (IFM) consists of the rest of the operations, which fuse source and target information into the final layer representation.
\end{itemize}
Compared with the standard splits~\cite{Vaswani:2017:NIPS}, we associate the ``Add\&Norm'' operation after encoder attention with the IFM, since it  starts the process of information fusion by a simple additive operation. Consequently, the functionalities of the three modules are well-separated.

\subsection{Research Questions}

Modern Transformer decoder is implemented as multiple identical layers, in which the source and target information are exploited and evolved layer-by-layer.
One research question arises naturally:
\begin{framed}
\noindent {\bf RQ1}. How do source and target information evolve within the decoder layer-by-layer and module-by-module?
\end{framed}
\noindent In Section~\ref{sec:information-quantify}, we introduce a universal probing scheme to quantify the amount of information embedded in decoder modules and explore their evolutionary trends.
The general trend we find is that higher layers contain more source and target information, while the sub-layers behave differently.
Specifically, the amount of information contained by SEMs would first increase and then decrease.
In addition, we establish that SEM guides both source and target information evolution within the decoder.

Since SEMs are critical to the decoder representation evolution, we conduct a more detailed study into the internal behaviors of the SEMs. The exploitation of source information is also closely related to the inadequate translation problem -- a key weakness of NMT models~\cite{Tu:2016:ACL}. We try to answer the following research question:
\begin{framed}
\noindent {\bf RQ2}. How does SEM exploit the source information in different layers?
\end{framed}
\noindent In Section~\ref{sec:source-exploitation}, we investigate how the SEMs transform the source information to the target side in terms of alignment accuracy and coverage ratio~\cite{Tu:2016:ACL}. Experimental results show that higher layers of SEM modules accomplish word alignment, while lower layer ones exploit necessary contexts.
This also explains the representation evolution of source information:
lower layers collect {\em more} source information to obtain a global view of source input, and higher layers extract {\em less} aligned source input for accurate translation.

Of the three sub-layers, IFM modules conceptually appear to play a key role in merging source and target information -- raising our final question:
\begin{framed}
\noindent {\bf RQ3}. How does IFM fuse source and target information on the operation level?
\end{framed}
\noindent In Section~\ref{sec:information-fusion}, we first conduct a fine-grained analysis of the IFM module on the operation level, and find that a simple ``Add\&Norm'' operation performs just as well at fusing information.
Thus, we {\em simplify} the IFM module to be only one Add\&Norm operation. Surprisingly, this performs similarly to the full model while significantly reducing the number of parameters and consequently boosting both training and inference speed.

\section{Experiments}

\paragraph{Data}

To make our conclusions compelling, all experiments and analysis are conducted on three representative language pairs.
For English$\Rightarrow$German (En$\Rightarrow$De), we use WMT14 dataset that consists of 4.5M sentence pairs.
The English$\Rightarrow$Chinese (En$\Rightarrow$Zh) task is conducted on WMT17 corpus, consisting of 20.6M sentence pairs.
For English$\Rightarrow$French (En$\Rightarrow$Fr) task, we use WMT14 dataset that comprises 35.5M sentence pairs.
English and French have many aspects in common while English and German differ in word order, requiring a significant amount of reordering in translation. Besides, Chinese belongs to a different language family compared to the others.

\paragraph{Models}

We conducted the experiments on the state-of-the-art Transformer~\cite{Vaswani:2017:NIPS}, and implemented our approach with the open-source toolkit FairSeq~\cite{ott2019fairseq}.
We follow the setting of Transformer-Base in ~\citet{Vaswani:2017:NIPS}, which consists of 6 stacked encoder/decoder layers with the model size being 512.
We train our models on 8 NVIDIA P40 GPUs, where each is allocated with a batch size of 4,096 tokens. We use Adam optimizer~\cite{kingma2014adam} with 4,000 warm-up steps.\footnote{More implementation details are in Sec~\ref{sec:app_implement}.}

\begin{figure}[t]
	\centering
	\includegraphics[width=0.48\textwidth]{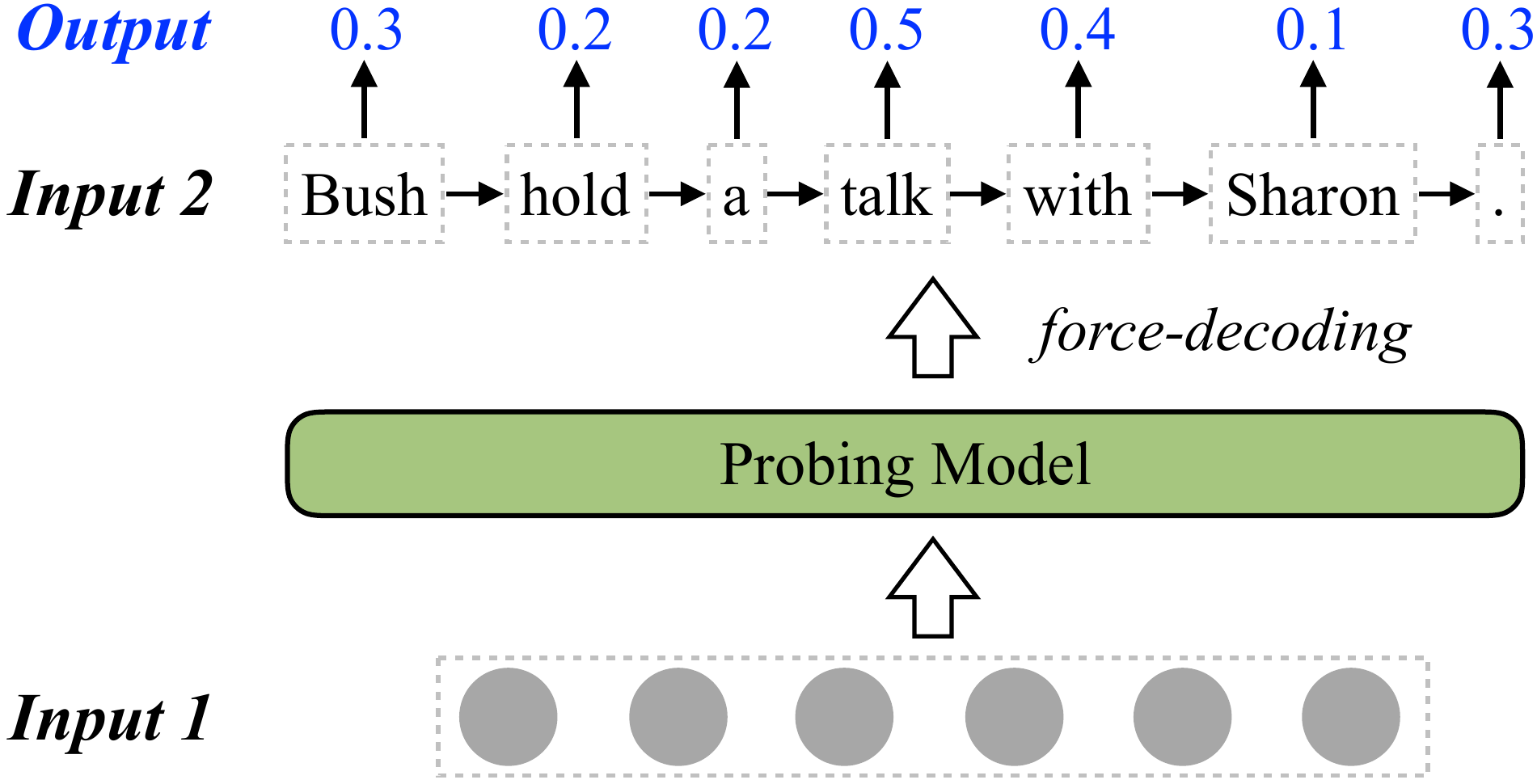}
	\caption{Illustration of the information probing model, which reads the representation of a decoder module (``Input 1'') and the word sequence to recover (``Input 2''), and outputs the generation probability (``Output'').}
	\label{fig:prob_arch}
\end{figure}

\subsection{Representation Evolution Across Layers}
\label{sec:information-quantify}

In order to quantify and visualize the representation evolution, we design a universal probing scheme to quantify the source (or target) information stored in network representations.

\paragraph{Task Description}
Intuitively, the more the source (or target) information stored in a network representation, the more probably a trained reconstructor could recover the source (or target) sequence.
Since the lengths of source sequence and decoder representations are not necessarily the same, the widely-used classification-based probing approaches~\cite{belinkov2017neural,tenney2019you} cannot be applied to this task. Accordingly, we cast this task as a generation problem -- evaluating the likelihood of generating the word sequence conditioned on the input representation.

Figure~\ref{fig:prob_arch} illustrates the architecture of our probing scheme.
Given a representation sequence from decoder ${\bf H} = \{{\bf h}_1, \dots, {\bf h}_M \}$ and the source (or target) word sequence to be recovered ${\bf x}=\{x_1, \dots, x_N\}$
the recovery likelihood is calculated as the perplexity (i.e. negative log-likelihood) of forced-decoding the word sequence:
\begin{equation}
    PPL({\bf x}|{\bf H}) = \sum_{n=1}^N -\log P(x_n | x_{<n}, {\bf H}) \label{eq:loss}
\end{equation}
The lower the recovery perplexity, the more the source (or target) information stored in the representation.
The probing model can be implemented as any architecture. For simplicity, we use a one-layer Transformer decoder.
{\em We train the probing model to recover both source and target sequence from all decoder sub-layer representations.}
During training, we fix the NMT model parameters and train the probing model on the MT training set to minimize the recovery perplexity in Equation~\ref{eq:loss}.

\paragraph{Task Discussion}
The above probing scheme is a general framework applicable to probing any given sequence from a network representation.
When we probe for the source sequence, the probing model is analogous to an auto-encoder~\cite{Bourlard:1988:BC,Vincent:2010:JLMR}, which reconstructs the original input from the network representations.
When we probe for the target sequence, we apply an attention mask to the {\it probing decoder} to avoid direct copying from the input of translation histories. Contrary to source probing, the target sequence is never seen by the model.

In addition, our proposed scheme can also be applied to probe linguistic properties that can be represented in a sequential format. For instance, we could probe source constituency parsing information, by training a probing model to recover the linearized parsing sequence~\cite{vinyals2015grammar}. Due to space limitations, we leave the linguistic probing to future work.

\begin{figure}[t]
	\centering
	\subfloat[En-De]{
		\includegraphics[height=0.265\textwidth]{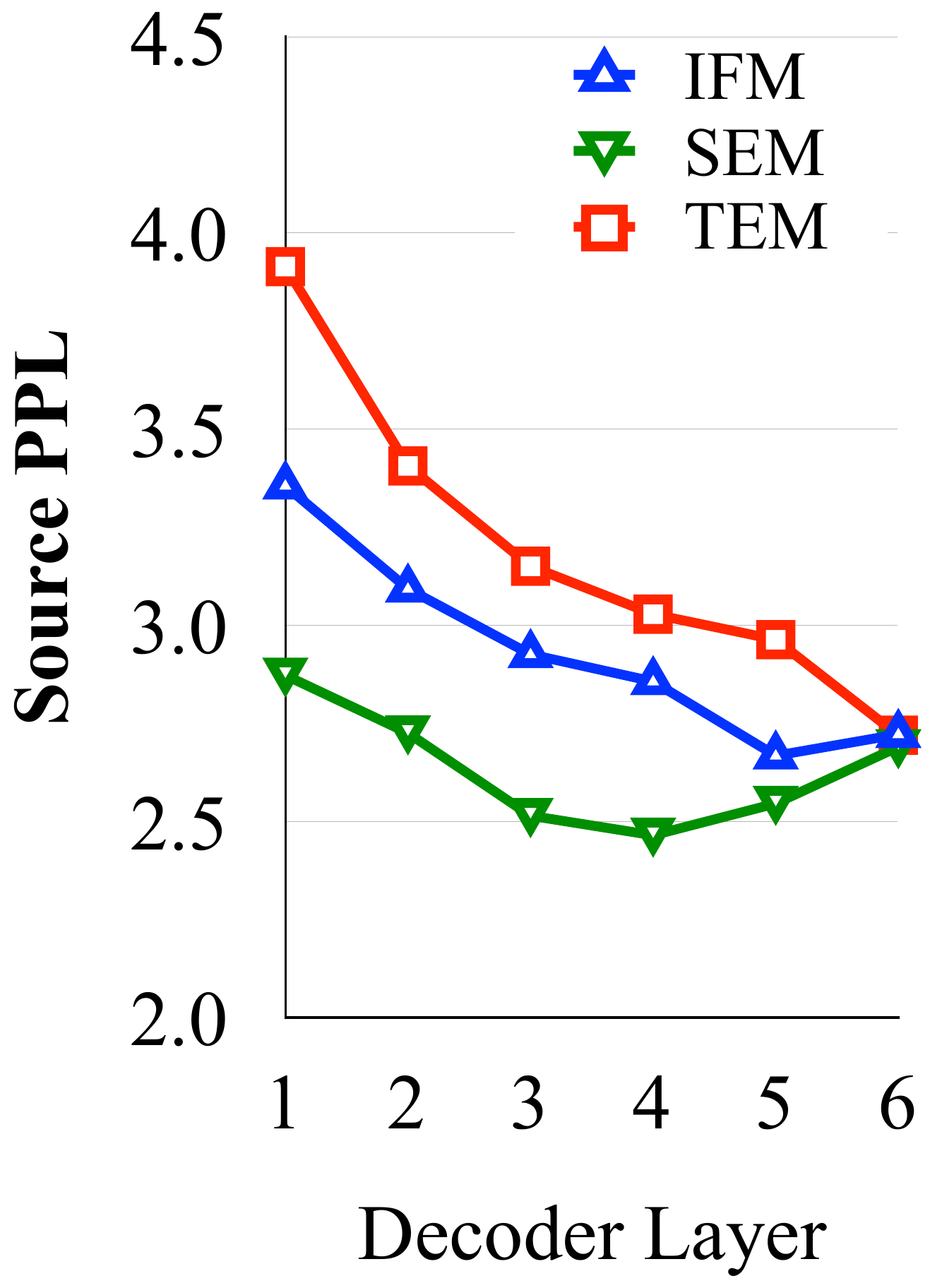}}
	\subfloat[En-Zh]{
		\includegraphics[height=0.265\textwidth]{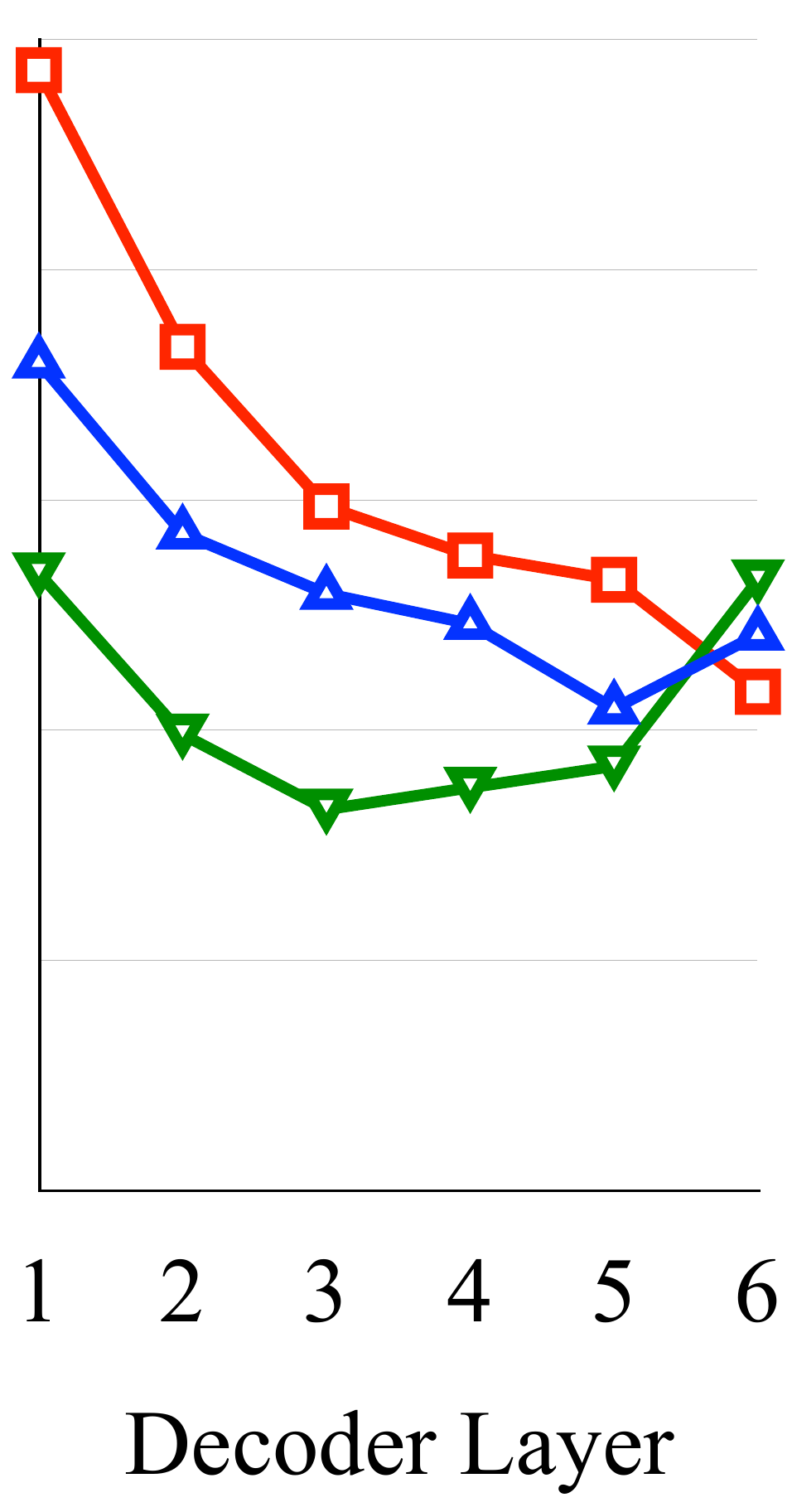}}
	\subfloat[En-Fr]{
		\includegraphics[height=0.265\textwidth]{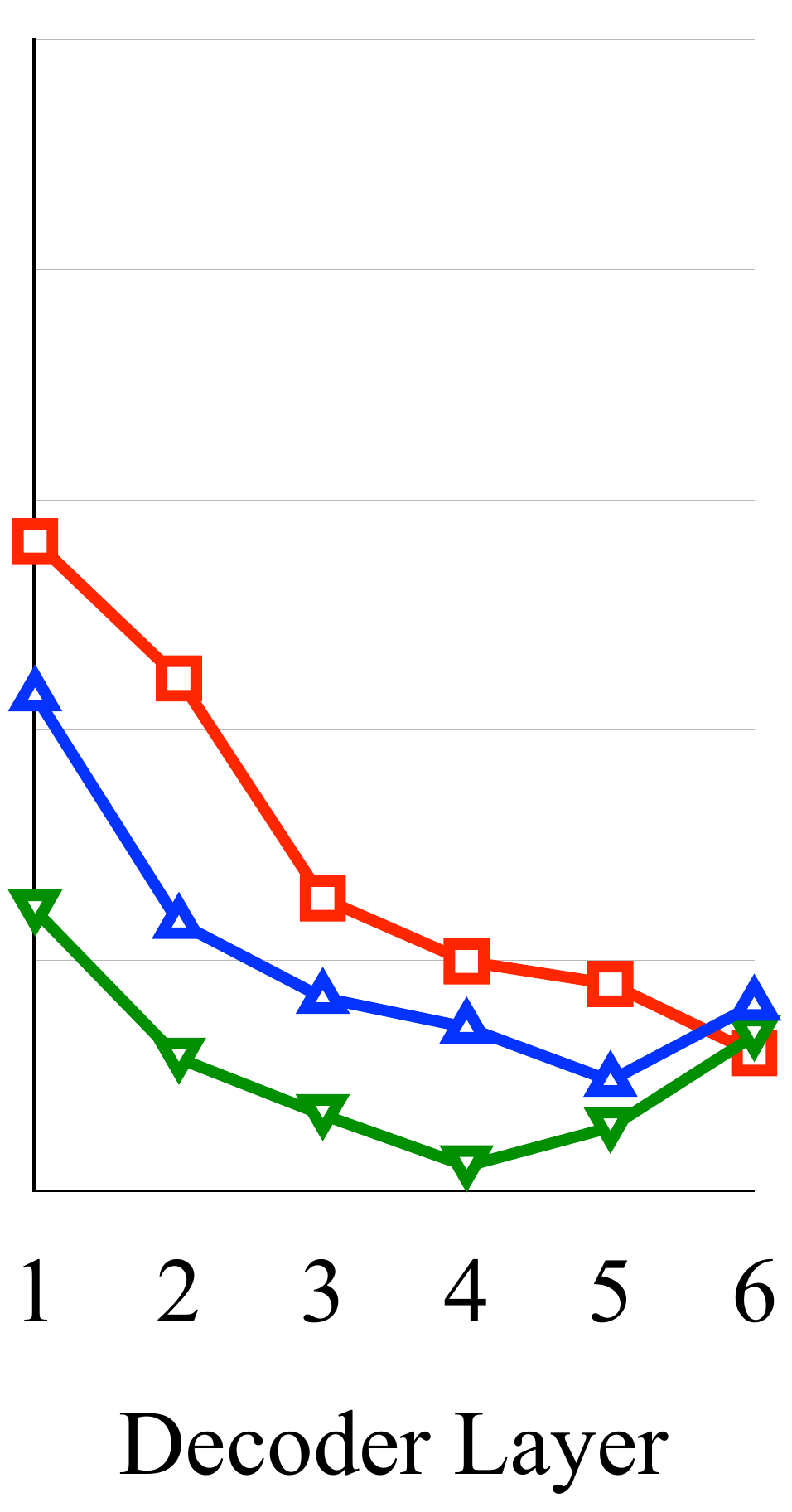}}
    \\
	\subfloat[En-De]{
		\includegraphics[height=0.265\textwidth]{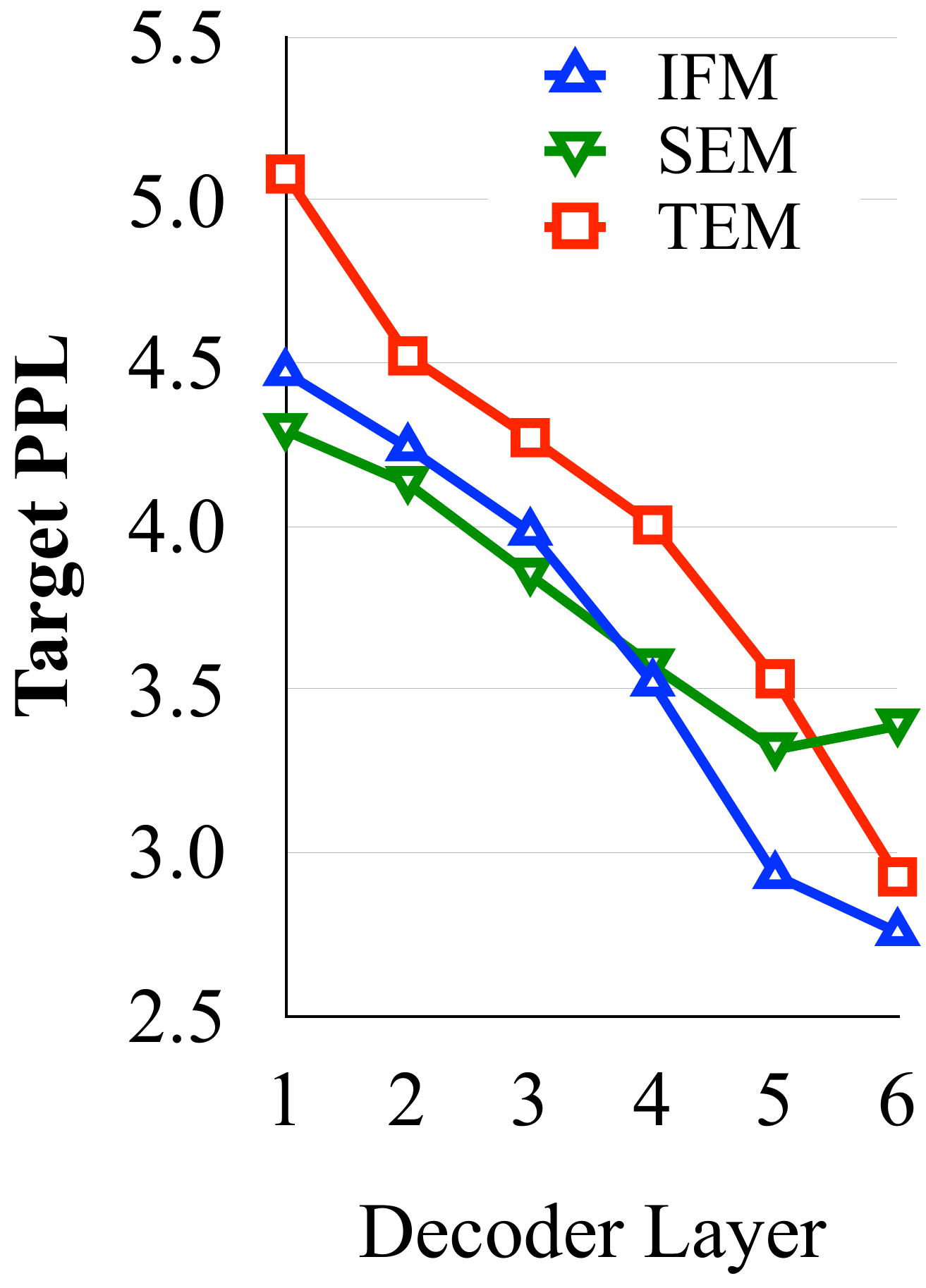}}
	\subfloat[En-Zh]{
		\includegraphics[height=0.265\textwidth]{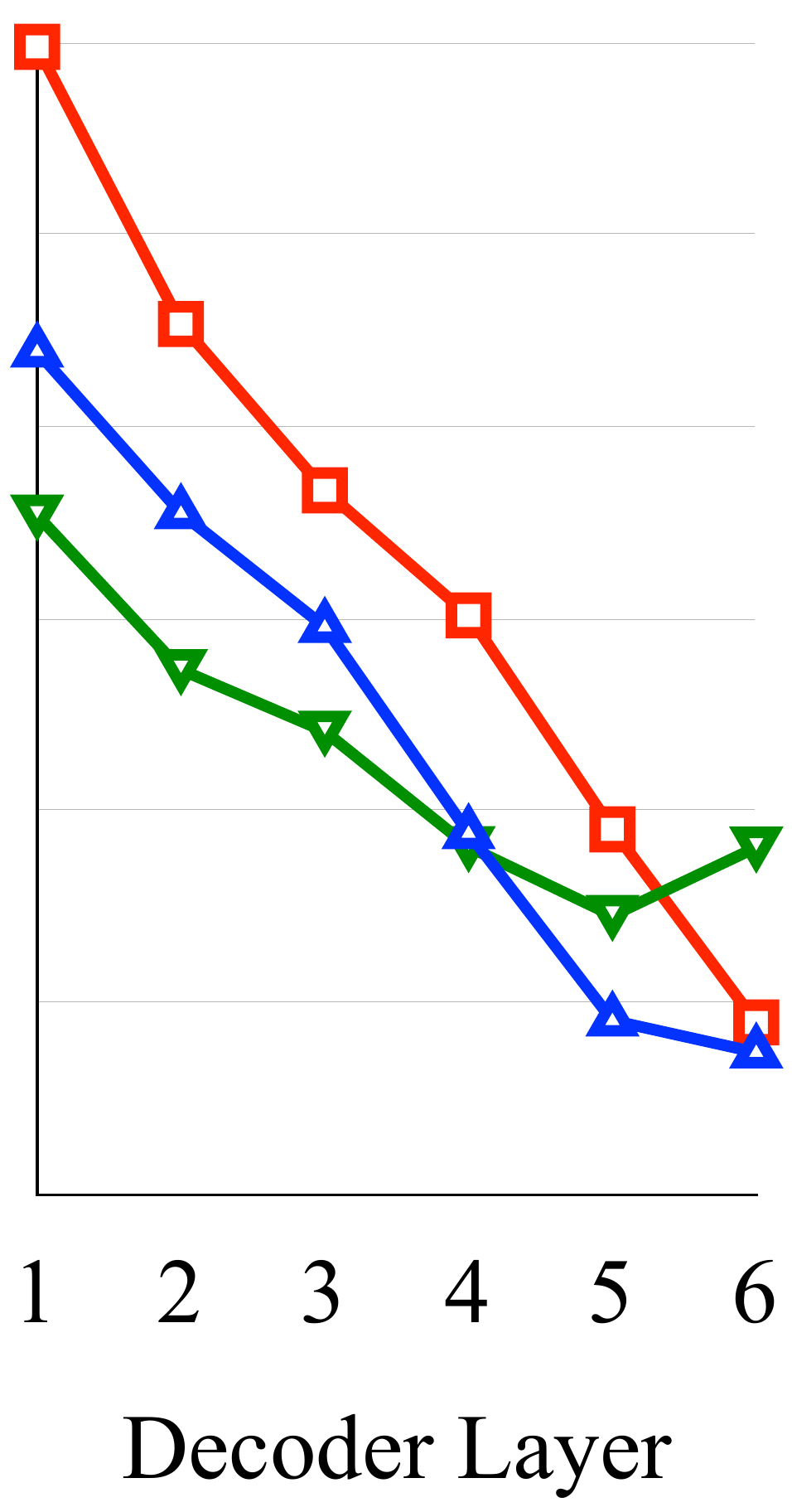}}
	\subfloat[En-Fr]{
		\includegraphics[height=0.265\textwidth]{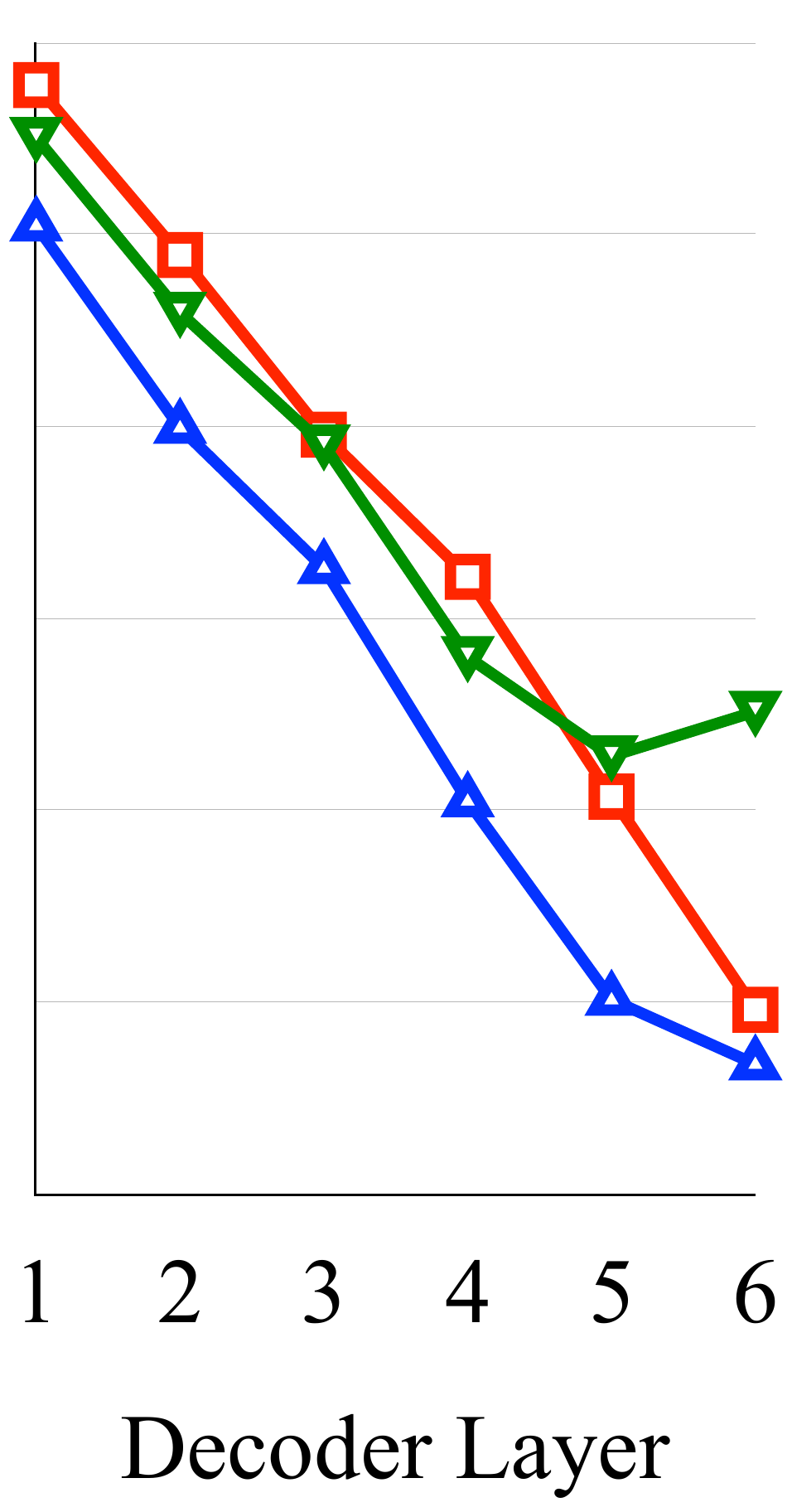}}
	\caption{Evolution trends of source (upper panel) and target (bottom panel) information embedded in the decoder modular representations across layers. Lower perplexity (``PPL'') denotes more information embedded in the representations.}
	\label{fig:layers}
\end{figure}

\paragraph{Probing Results}
Figure~\ref{fig:layers} shows the results of our information probing conducted on the heldout set.
We have a few observations:
\begin{itemize}
    \item The evolution trends of TEM and IFM are largely the same. Specifically, the curve of TEM is very close to that of IFM shifted up by one layer. Since TEM representations are two operations (self-attn. and Add\&Norm) away from the previous layer IFM, this observation indicates TEMs do not significantly affect the amount of source/target information.
    \footnote{
    TEM may change the order or distribution of source/target information, which are not captured by our probing experiments.}
    \item {\em SEM guides both source and target information evolution}.
    While closely observing the curves, the trend of layer representations (i.e. IFM) is always led by that of SEM.
    For example, as the PPL of SEM transitions from decreases to increases, the PPL of IFM slows down the decreases and starts increasing as an aftermath.
    This is intuitive: in machine translation, source and target sequences should contain equivalent information, thus the target generation should largely follow the lead of source information (from SEM representations) to guarantee its adequacy.
    \item For IFM,
    the amount of target information consistently increases in higher layers -- a consistent decrease of PPL in Figures~\ref{fig:layers}(d-f).
    While source information goes up in the lower layers, it drops in the highest layer (Figures~\ref{fig:layers}(a-c)).
\end{itemize}
Since SEM representations are critical to decoder evolution, we turn to investigate how SEM exploit source information, in the hope of explaining the decoder information evolution.

\subsection{Exploitation of Source Information}
\label{sec:source-exploitation}

\begin{figure}[t]
	\centering
	\subfloat[Word Alignment]{
		\includegraphics[height=0.3\textwidth]{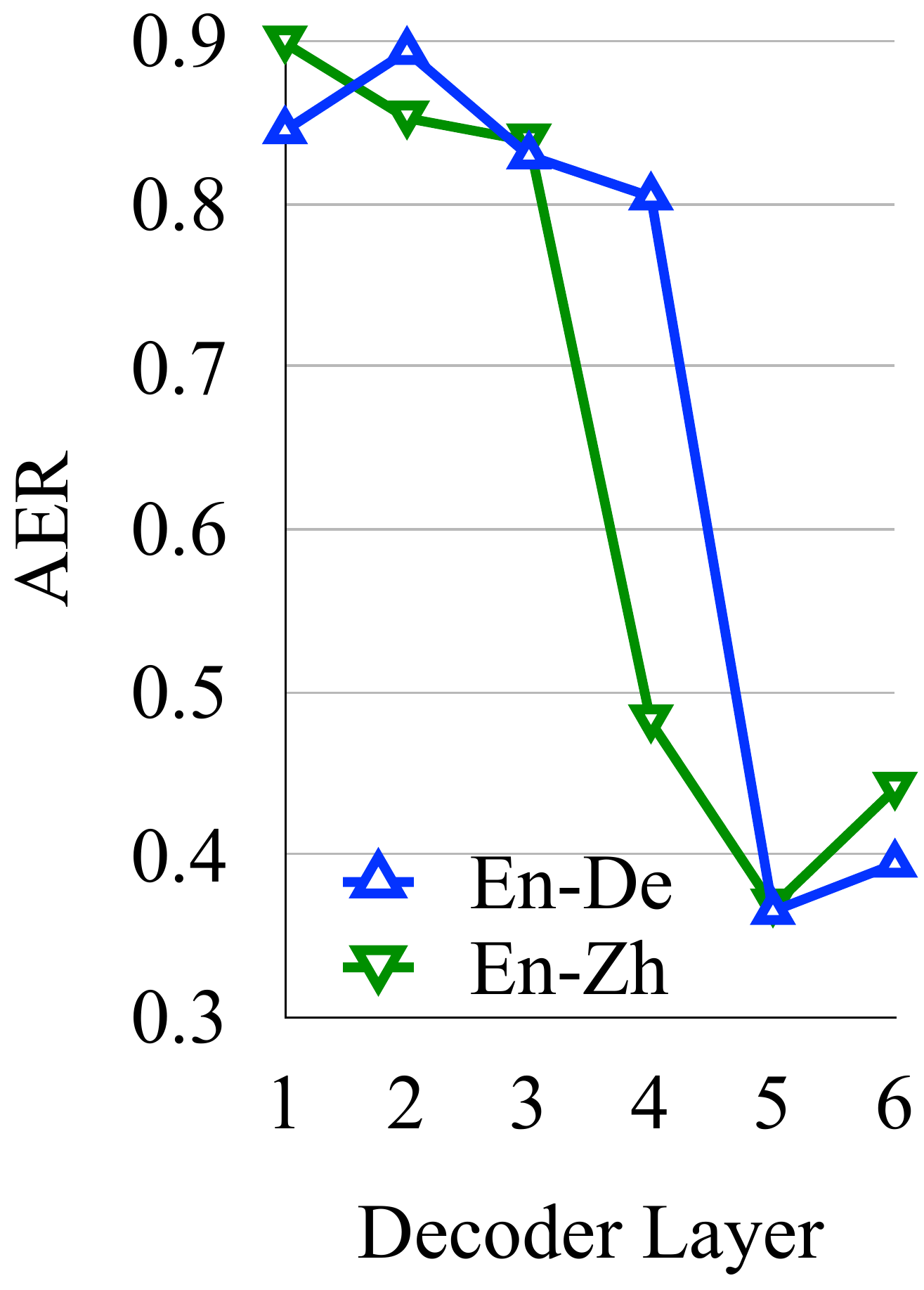}}
	\hfill
	\subfloat[Cumulative Coverage]{
		\includegraphics[height=0.3\textwidth]{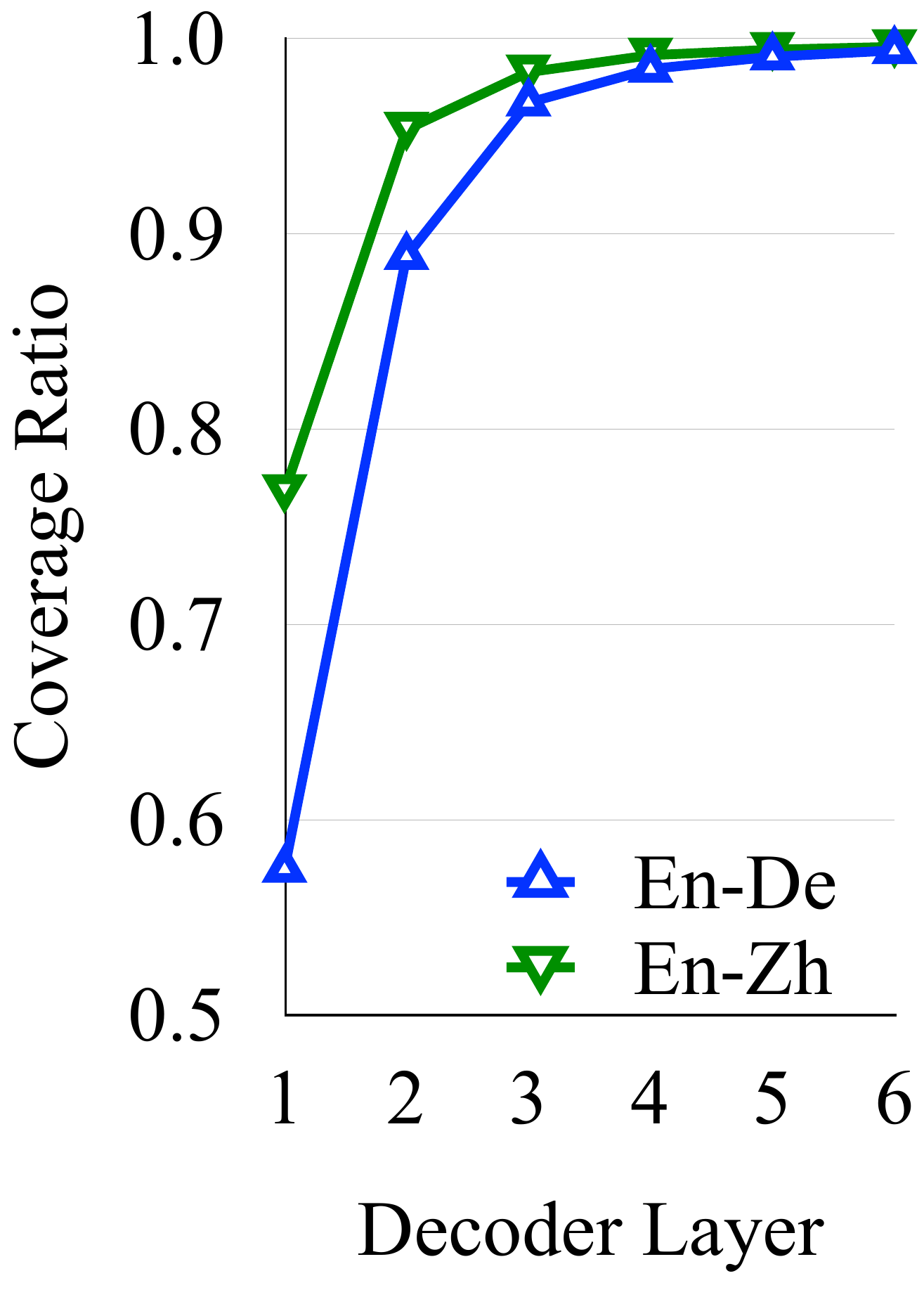}}
	\caption{Behavior of the SEM in terms of (a) alignment quality measured in AER (the lower, the better), and (b)  the cumulative coverage of source words.}
	\label{fig:SEM}
\end{figure}

Ideally, SEM should {\em accurately} and {\em fully} incorporate the source information for the decoder. Accordingly, we evaluate how well SEMs accomplish the expected functionality from two perspectives.

\paragraph{\em Word Alignment.} Previous studies generally interpret the attention weights of SEM as word alignments between source and target words, which can measure whether SEMs select the most relevant part of source information for each target token~\cite{Tu:2016:ACL,Li:2019:ACL,Tang:2019:RANLP}.
We follow previous practice to merge attention weights from the SEM attention heads, and to extract word alignments by selecting the source word with the highest attention weight for each target word. We calculate the alignment error rate (AER) scores~\cite{och2003systematic} for word alignments extracted from SEM of each decoder layer.

\paragraph{\em Cumulative Coverage.}
Coverage is commonly used to evaluate whether the source words are fully translated~\cite{Tu:2016:ACL,Kong:2019:AAAI}. We use the above extracted word alignments to identify the set of source words $A_i$, which are covered (i.e., aligned to at least one target word) at each layer. We then propose a new metric {\em cumulative coverage ratio} $C_{\leq i}$ to indicate how many source words are covered by the layers $\leq i$:
\begin{equation}
    C_{\leq i} = \frac{|A_1 \cup \dots \cup A_i|}{N}
\end{equation}
where $N$ is the number of total source words. This metric indicates the completeness of source information coverage till layer $i$.

\paragraph{Dataset}
We conducted experiments on two manually-labeled alignment datasets: RWTH En-De\footnote{\url{https://www-i6.informatik.rwth-aachen.de/goldAlignment}} and En-Zh~\cite{liu2015contrastive}. The alignments are extracted from NMT models trained on the WMT En-De and En-Zh dataset.

\begin{figure}[t]
	\centering
	\subfloat[Alignment]{
		\includegraphics[height=0.3\textwidth]{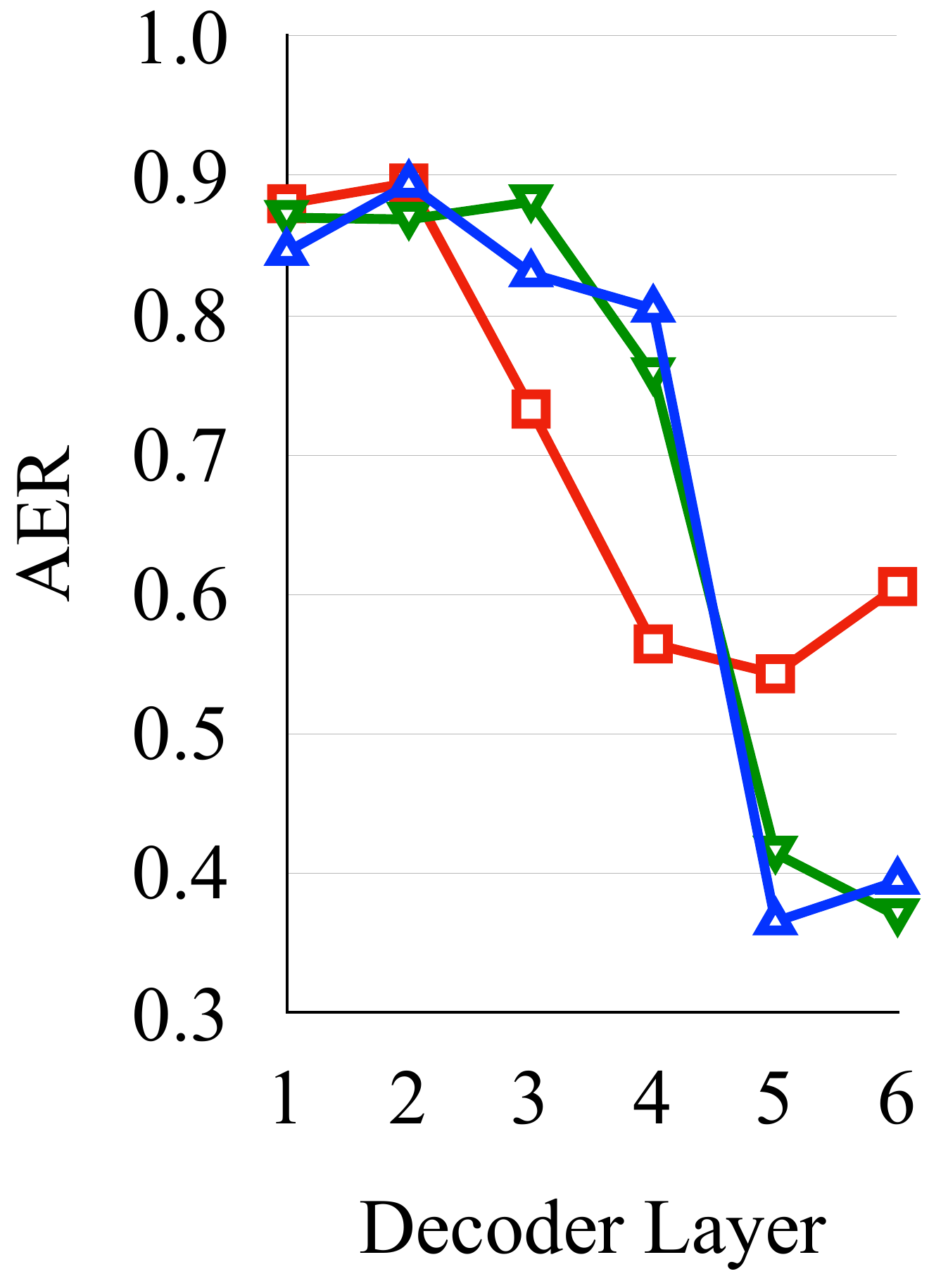}}
	\hfill
	\subfloat[Cumulative Coverage]{
		\includegraphics[height=0.3\textwidth]{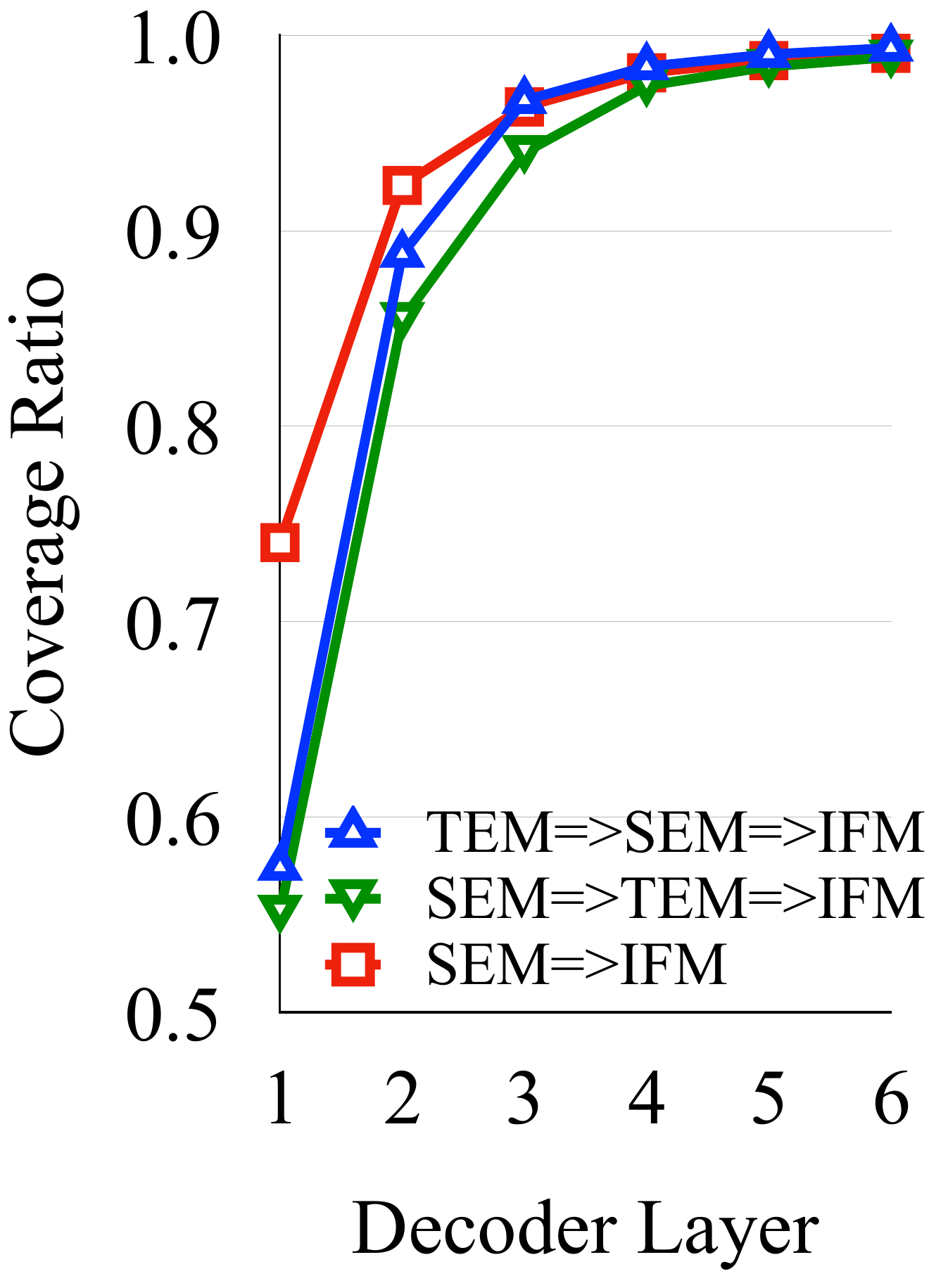}}
	\caption{Effects of the stacking order of TEM and SEM on the En-De dataset.}
	\label{fig:order}
\end{figure}

\begin{figure}[t]
	\centering
	\subfloat[Alignment]{
		\includegraphics[height=0.3\textwidth]{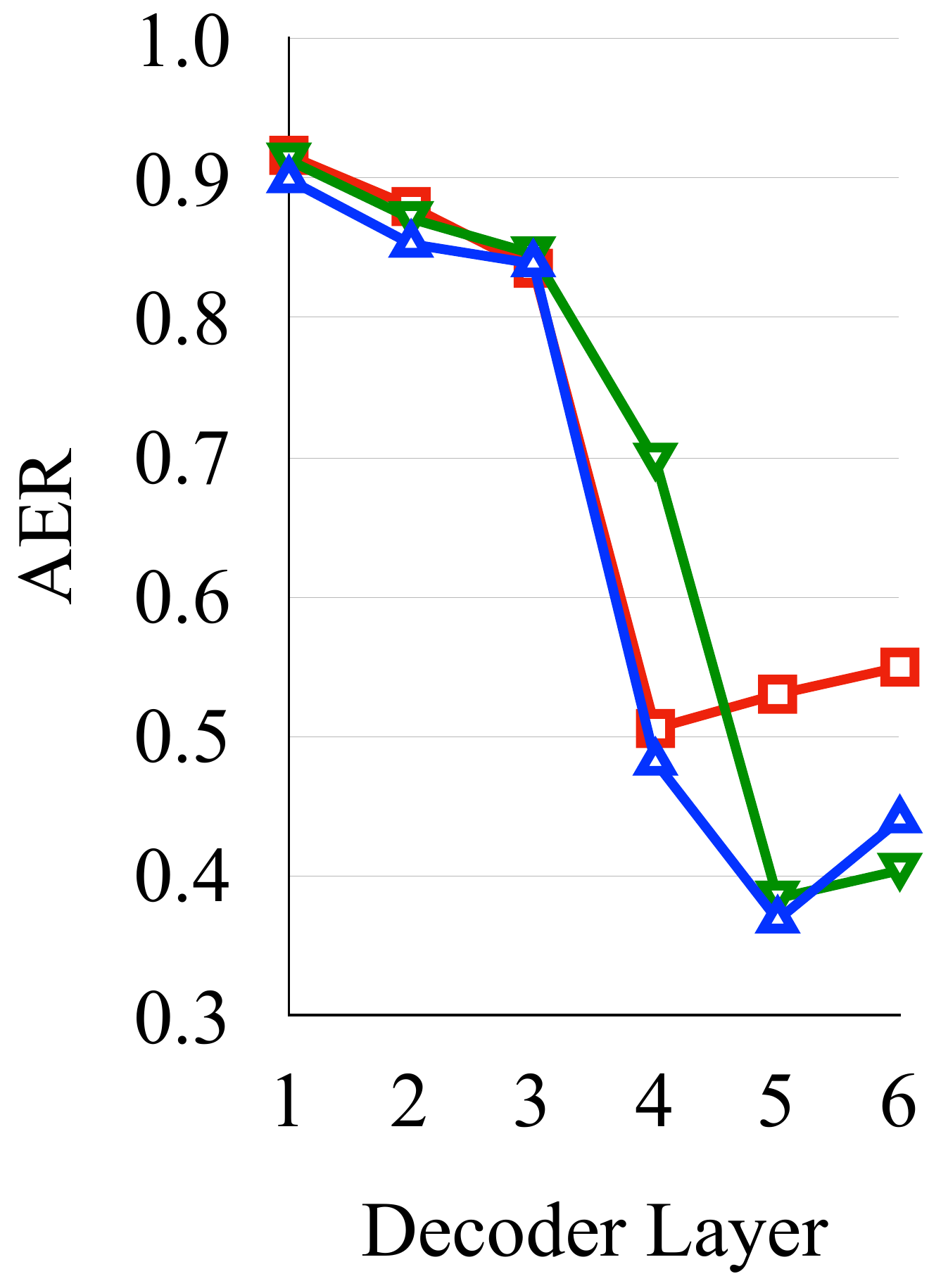}}
	\hfill
	\subfloat[Cumulative Coverage]{
		\includegraphics[height=0.3\textwidth]{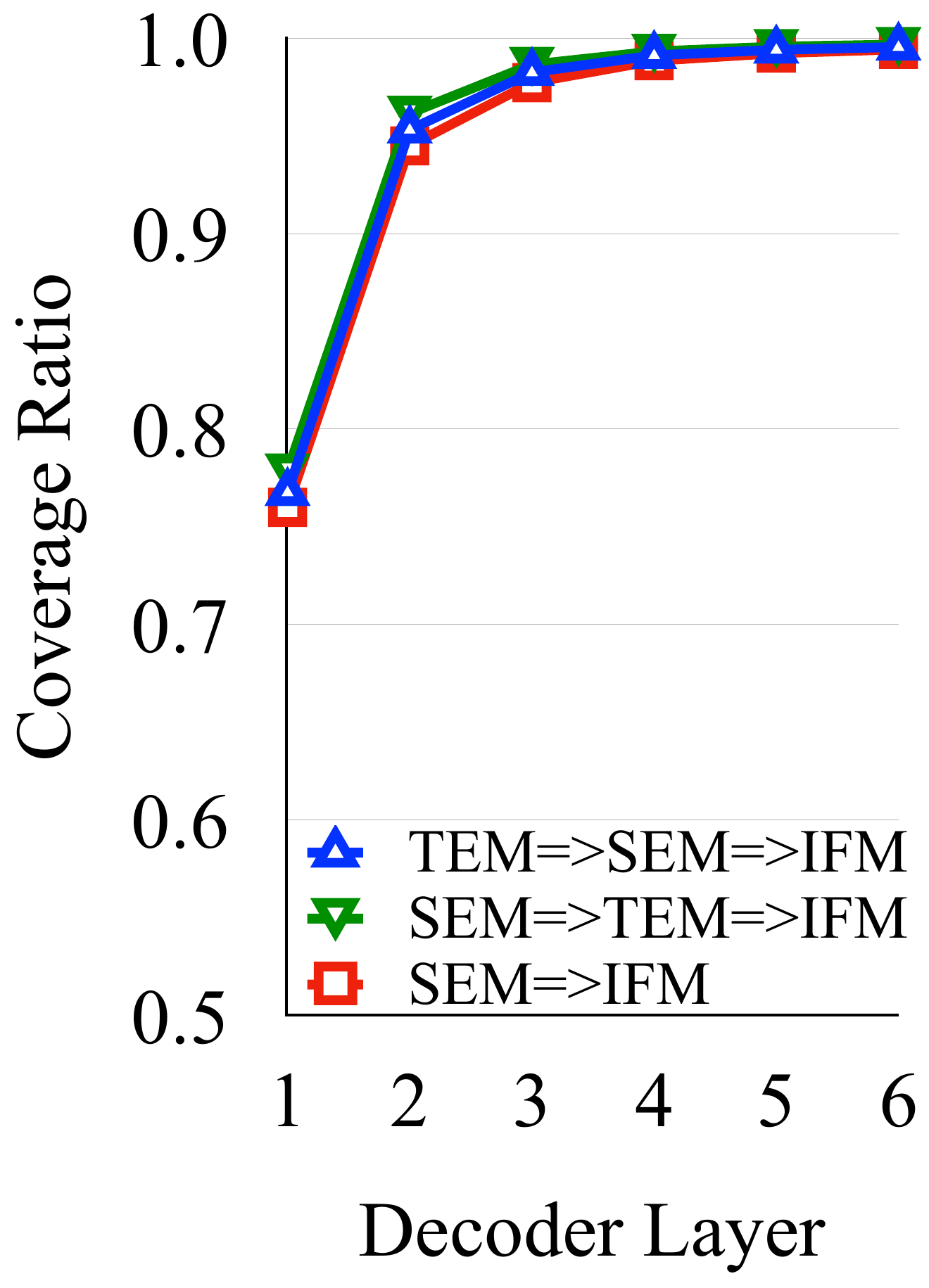}}
	\caption{Effects of the stacking order of TEM and SEM on the En-Zh dataset.}
	\label{fig:order_enzh}
\end{figure}

\paragraph{Results}
Figure~\ref{fig:SEM} demonstrates our results on word alignment and cumulative coverage.
We find that the {\em lower-layer SEMs focus on gathering source contexts} (rapid increase of cumulative coverage with poor word alignment), while {\em higher-layer ones play the role of word alignment} with the lowest AER score of less than 0.4 at the 5th layer.
The $4^{th}$ layer and the $3^{rd}$ layer separate the two roles for En-De and En-Zh respectively. Correspondingly, they are also the turning points (PPL from decreases to increases) of source information evolution in Figure~\ref{fig:layers} (a,b).
Together with conclusions from Sec.~\ref{sec:information-quantify}, we demonstrate the general pattern of SEM: SEM tends to cover more source content and gain increasing amount of source information up to a turning point of $3^{rd}$ or $4^{th}$ layer, after which it starts only attending to the most relevant source tokens and contains decreasing amount of total source information.

\begin{table}[t]
	\centering
	\begin{tabular}{r||ccc}
		\bf Decoder                                                    & \bf En-De & \bf En-Zh & \bf En-Fr \\
		\hline
		\small \textsc{TEM}$\Rightarrow$\textsc{SEM}$\Rightarrow$\textsc{IFM} & 27.45 & 32.24  & 40.39 \\ \hline
		\small \textsc{SEM}$\Rightarrow$\textsc{TEM}$\Rightarrow$\textsc{IFM} & 27.61 & 33.62  & 40.89 \\
		\hdashline
		\small \textsc{SEM}$\Rightarrow$\textsc{IFM}                          & 22.76 & 30.06  & 37.56 \\
	\end{tabular}
	\caption{Effects of the stacking order of decoder sublayers on translation quality in terms of BLEU score.}
	\label{tab:order}
\end{table}

\paragraph{TEM Modules}

Since TEM representations serve as the query vector for encoder attention operations (shown in Figure~\ref{fig:architecture}), we naturally hypothesize that TEM is helping SEM on building alignments.

To verify that, we remove TEM from the decoder (``SEM$\Rightarrow$IFM''), which significantly increases the alignment error from 0.37 to 0.54 (in Figure~\ref{fig:order}), and leads to a serious decrease of translation performance (BLEU: 27.45 $\Rightarrow$ 22.76, in Table~\ref{tab:order}) on En-De, while results on En-Zh also confirms it (in Figure~\ref{fig:order_enzh}).
{\em This indicates that TEM is essential for building word alignment.}

However, reordering the stacking of TEM and SEM (``SEM$\Rightarrow$TEM$\Rightarrow$IFM'') does not affect the alignment or translation qualities (BLEU: 27.45 vs. 27.61). These results provide empirical support for recent work on merging TEM and SEM modules~\cite{Zhang:2019:EMNLP}.

\begin{table}[t]
	\centering
	\begin{tabular}{c||ccc|c}
		\bf Depth & \bf En-De & \bf En-Zh & \bf En-Fr & \bf Ave. \\
		\hline
		6         & 27.45     & 32.24     & 40.39     & 33.36  \\
		\hline
		4         & 27.52     & 31.35     & 40.37     & 33.08  \\
		12        & 27.64     & 32.50     & 40.44     & 33.53  \\
	\end{tabular}
	\caption{Effects of various decoder depths on translation quality in terms of BLEU score.}
	\label{tab:depth}
\end{table}

\begin{figure}[t]
	\centering
	\subfloat[Word Alignment]{
		\includegraphics[height=0.3\textwidth]{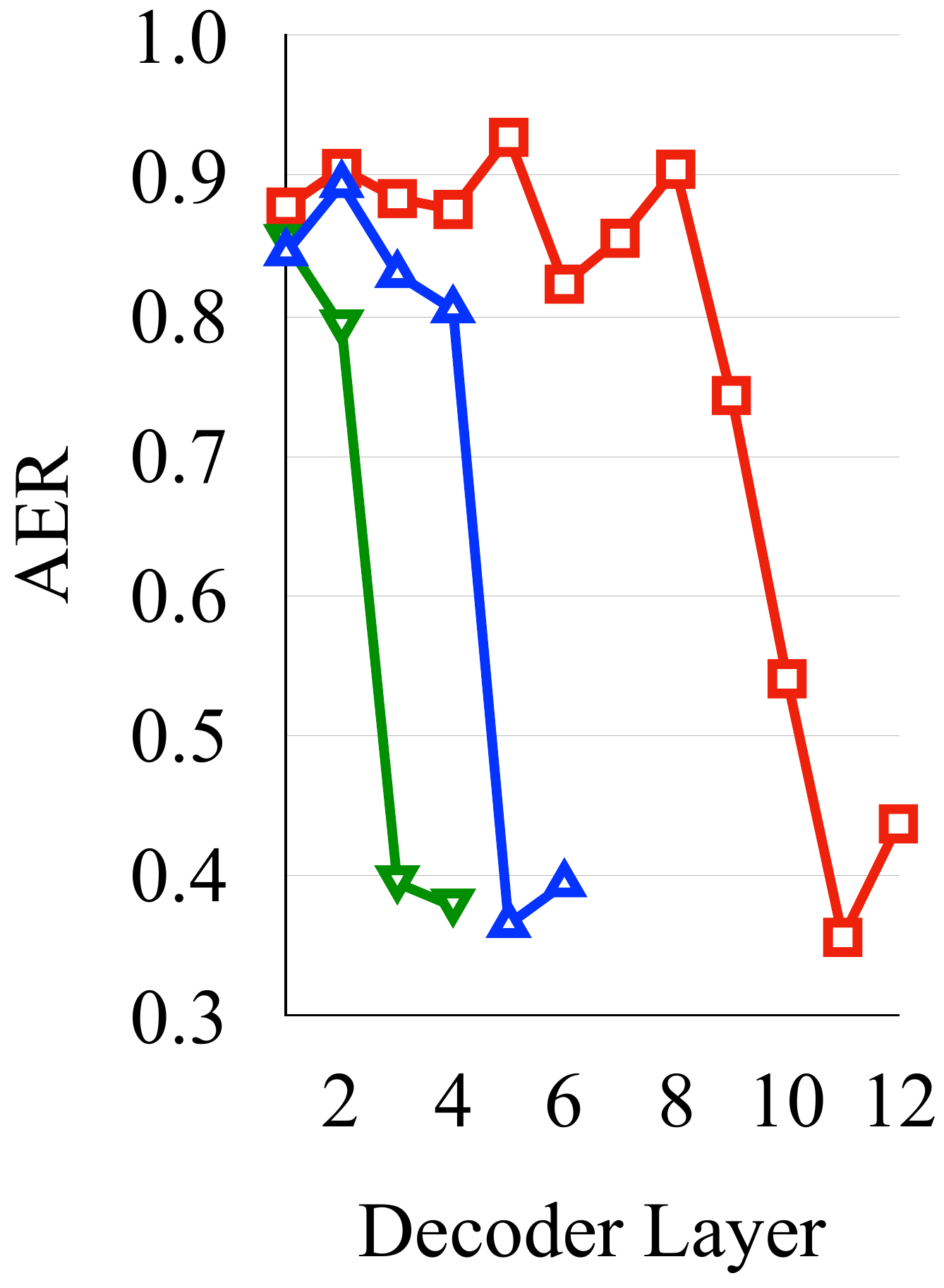}}
	\hfill
	\subfloat[Cumulative Coverage]{
		\includegraphics[height=0.3\textwidth]{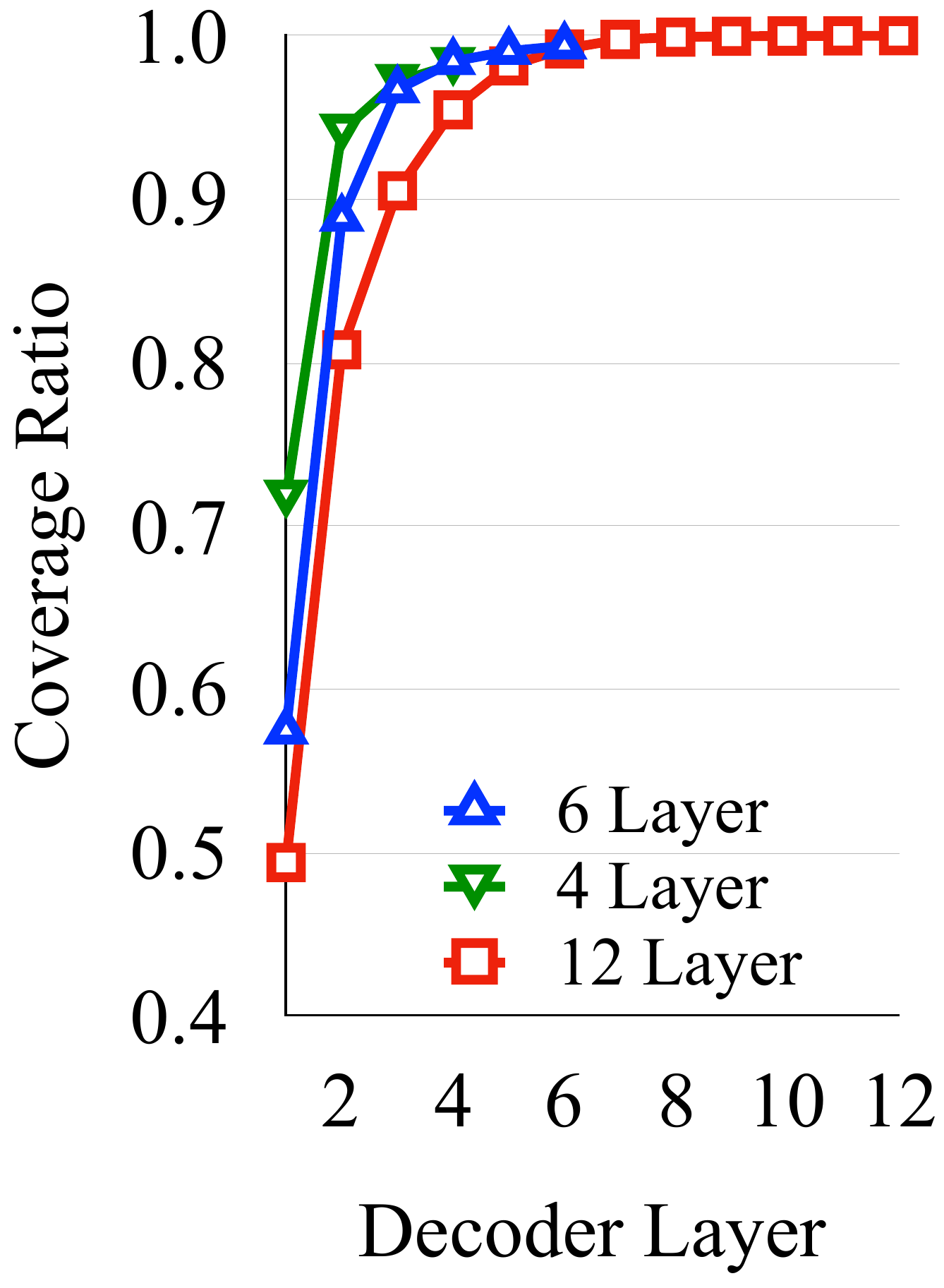}}
	\caption{Effects of decoder depths on SEM behaviors on the En-De task.}
	\label{fig:depth}
\end{figure}

\paragraph{Robustness to Decoder Depth}
To verify the robustness of our conclusions, we vary the depth of NMT decoder and train it from scratch.
Table~\ref{tab:depth} demonstrates the results on translation quality, which generally show that more decoder layers bring better performance.
Figure~\ref{fig:depth} shows that SEM behaves similarly regardless of depth.
These results demonstrate the robustness of our conclusions. 

\subsection{Information Fusion in Decoder}
\label{sec:information-fusion}

We now turn to the analysis of IFM.
Within the Transformer decoder, IFM plays the critical role of fusing the source and target information by merging representations from SEM and TEM.
To study the information fusion process, we conduct a more fine-grained analysis on IFM at the operation level.

\begin{figure}[t]
	\centering
	\subfloat[Three Operations in IFM]{\includegraphics[width=0.3\textwidth]{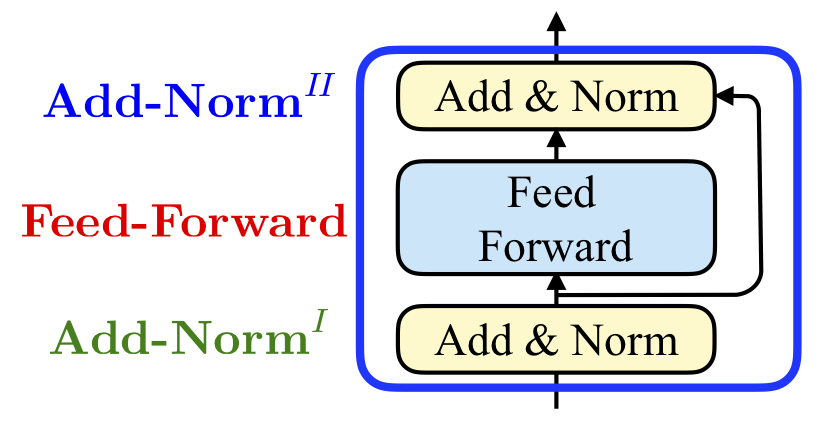}} \\
	\subfloat[Source PPL]{
		\includegraphics[height=0.3\textwidth]{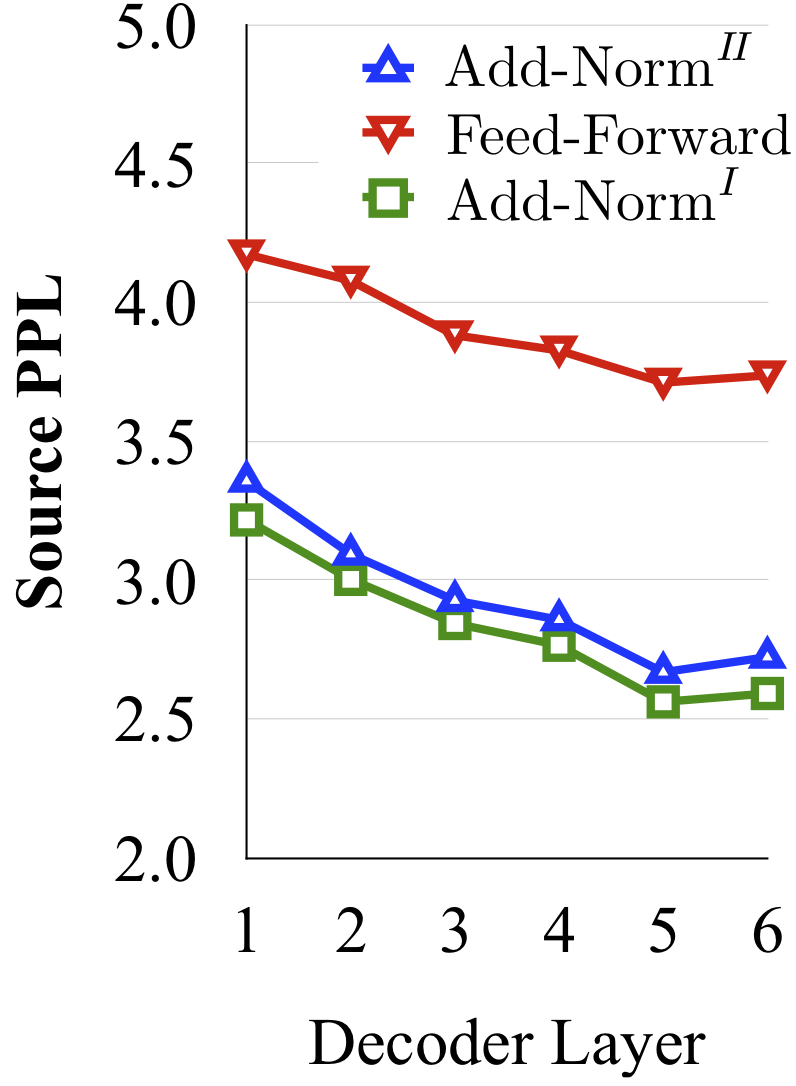}}
	\hfill
	\subfloat[Target PPL]{
		\includegraphics[height=0.3\textwidth]{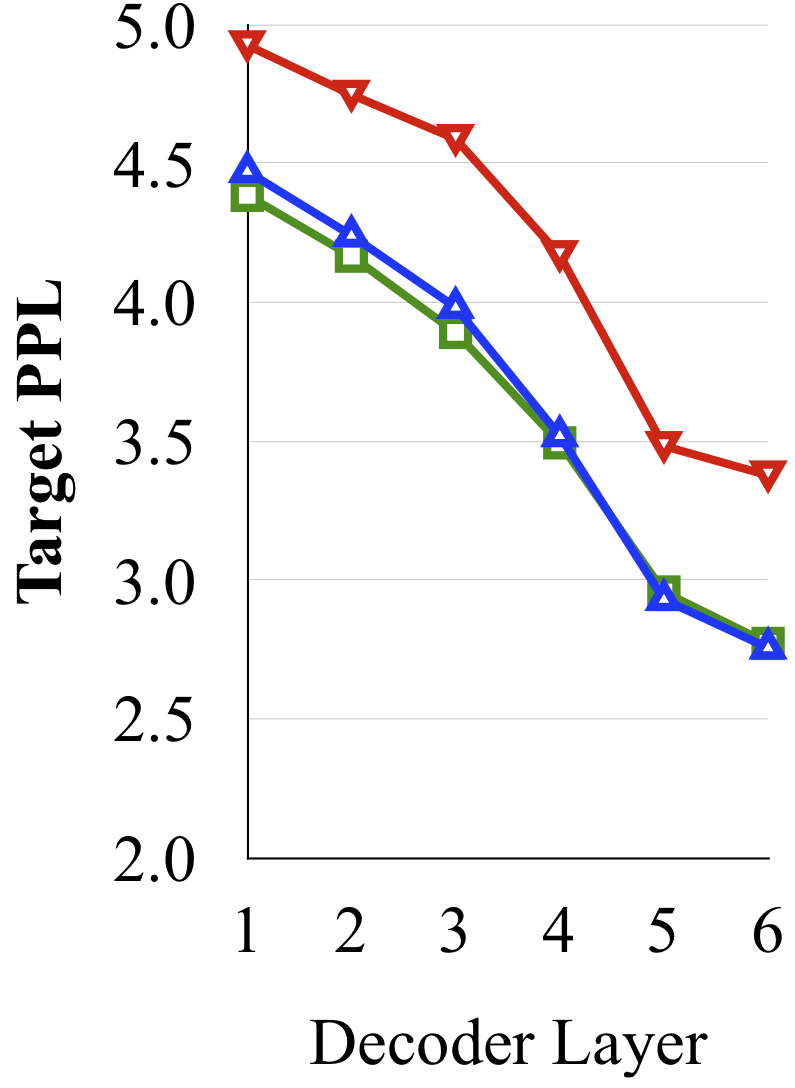}}
	\caption{Illustration of (a) three operations within IFM, and (b,c) the source and target information evolution within IFM on En-De task.}
	\label{fig:IFM_prob}
\end{figure}

\paragraph{Fine-Grained Analysis on IFM}
As shown in Figure~\ref{fig:IFM_prob}(a), IFM contains three operations:
\begin{compactitem}
    \item {\em Add-Norm$^I$} linearly sums and normalizes the representations from SEM and TEM;
    \item {\em Feed-Forward} non-linearly transforms the fused source and target representations;
    \item {\em Add-Norm$^{I\!I}$} again linearly sums and normalizes the representations from the above two.
\end{compactitem}
\paragraph{IFM Analysis Results}
Figures~\ref{fig:IFM_prob} (b) and (c) respectively illustrate the source and target information evolution within IFM.

Surprisingly, Add-Norm$^I$ contains a similar amount of, if not more, source (and target) information than Add-Norm$^{I\!I}$, while the Feed-Forward curve deviates significantly from both.
This indicates that the residual Feed-Forward operation may not affect the source (and target) information evolution, and one Add\&Norm operation may be sufficient for information fusion.

\begin{table}[t]
	\centering
	\begin{tabular}{c||ccc}
		{\bf Model} & {\bf Self-Attn.} & {\bf Enc-Attn.} & {\bf FFN} \\
		\hline
		Base   &  6.3M   &   6.3M   &  12.6M   \\
		\hline
		Big    &  25.2M    &   25.2M   &  50.4M   \\
	\end{tabular}
	\caption{Number of parameters taken by three major operations within Transformer Base and Big decoder.\footnotemark}
	\label{tab:parameters}
\end{table}

\paragraph{Simplified Decoder}

\begin{figure}[t]
	\centering
	\subfloat[Standard]{\includegraphics[width=0.2\textwidth]{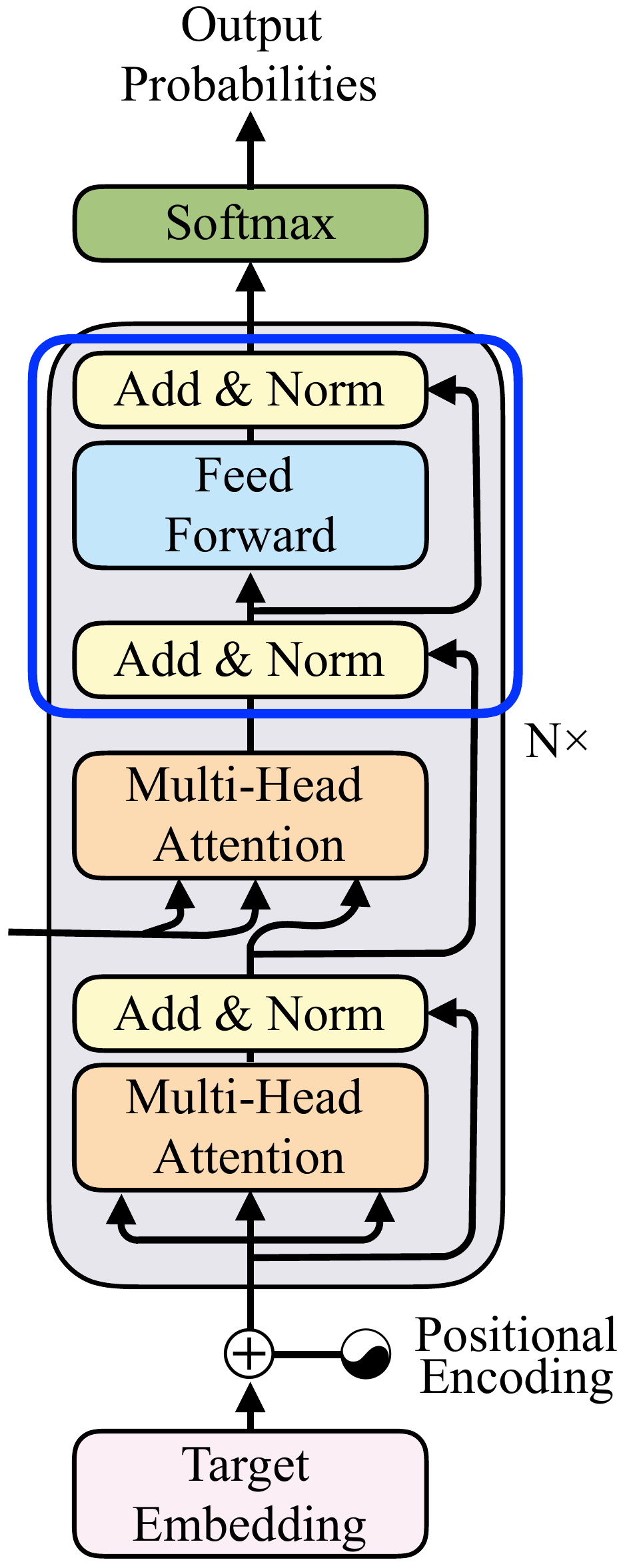}} \hspace{0.05\textwidth}
	\subfloat[Simplified]{\includegraphics[width=0.2\textwidth]{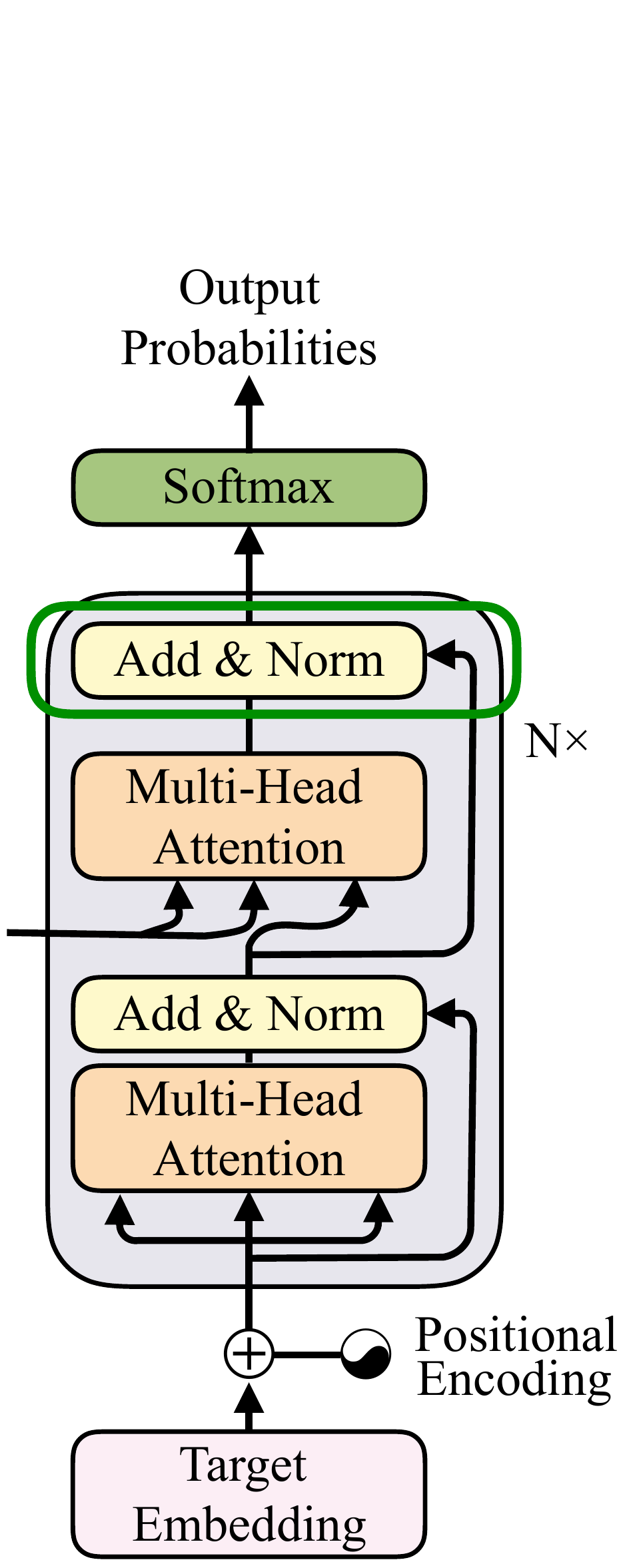}}
	\caption{Illustration of (a) the standard decoder, and (b) the simplified decoder with simplified IFM.}
	\label{fig:new_decoder}
\end{figure}

\begin{figure}[t]
	\subfloat[Source PPL]{
		\includegraphics[height=0.3\textwidth]{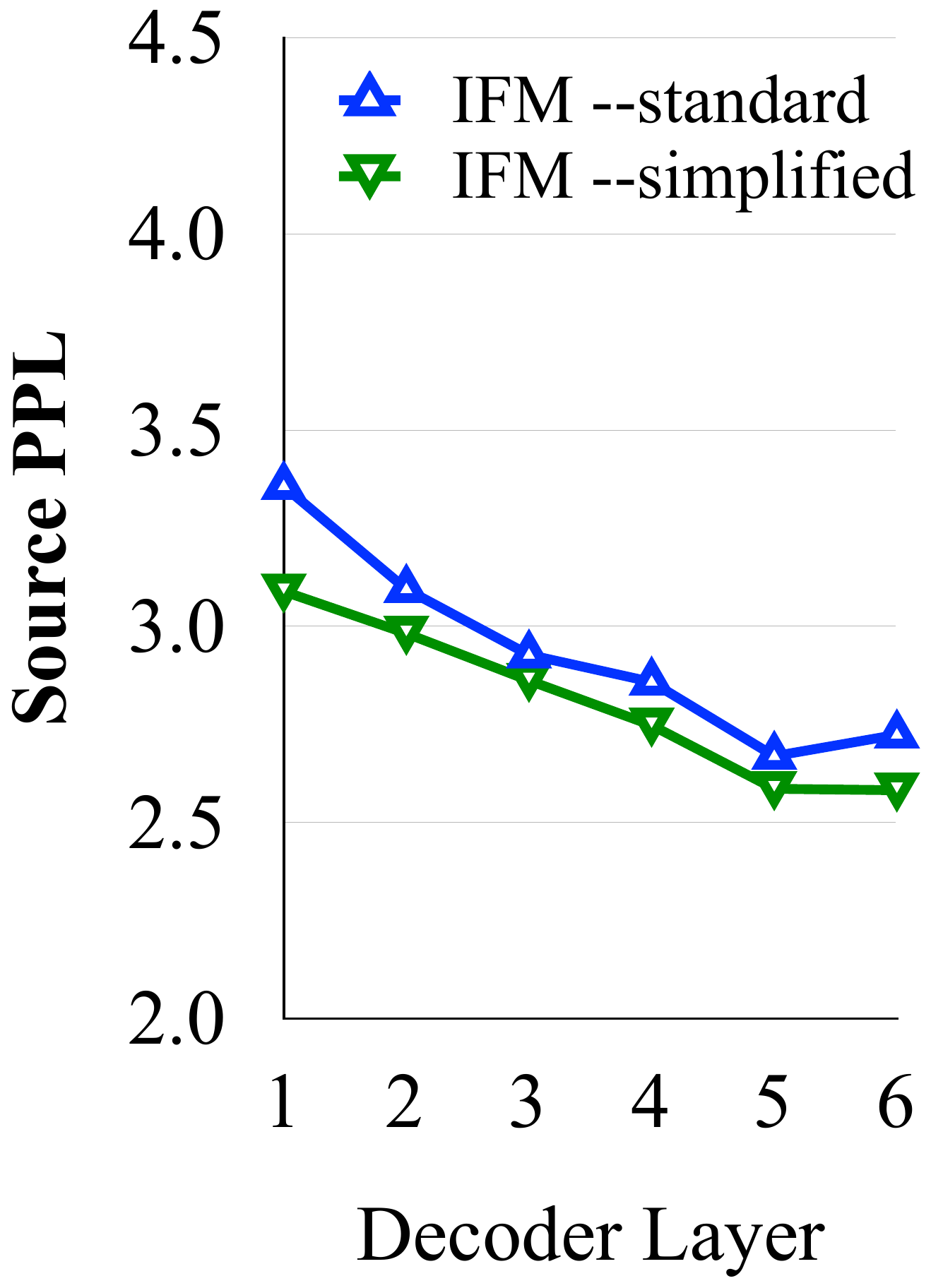}}
	\hfill
	\subfloat[Target PPL]{
		\includegraphics[height=0.3\textwidth]{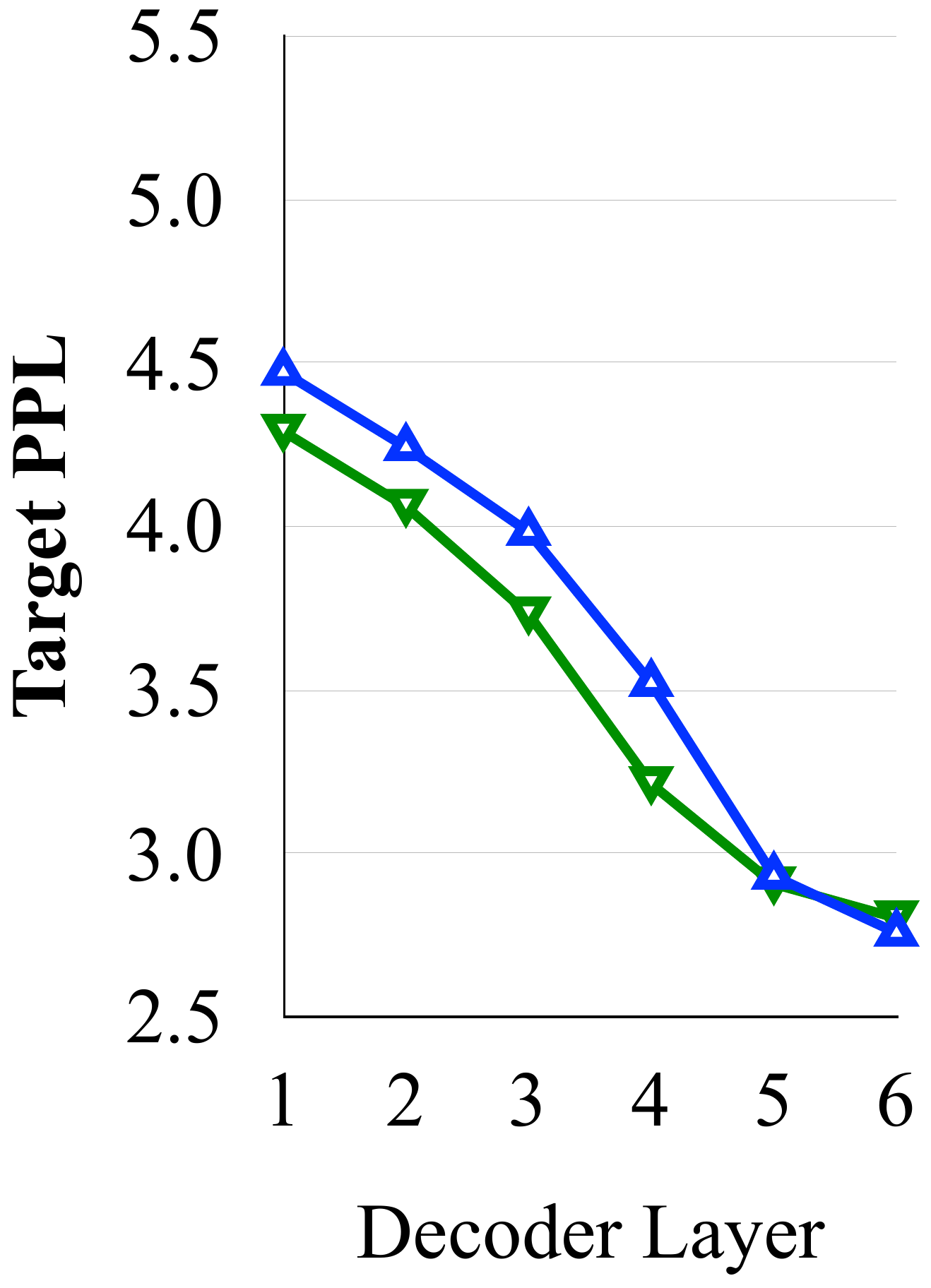}}
	\caption{Comparison of IFM information evolution between the standard and simplified decoder on En-De.}
	\label{fig:FFN_prob}
\end{figure}

To empirically demonstrate whether one Add\&Norm operation is already sufficient, we remove all other operations, leaving just one Add\&Norm operation for the IFM.
The architectural change is illustrated in Figure~\ref{fig:new_decoder}(b), and we dub it the ``{\em simplified decoder}''.

\begin{table}[t]
	\centering
	\begin{tabular}{c|c|rrr}
		\multicolumn{2}{c|}{\bf Decoder} & \bf BLEU & \bf \#Train. & \bf \#Infer. \\
		\hline
		\multirow{3}{*}{\rotatebox[origin=c]{90}{\bf En-De}}
		& Standard   & 27.45 & 63.93K & 65.35 \\
        & Simplified & 27.29 & 71.08K & 72.93 \\
        \cdashline{2-5}
        & $\triangle$ & \em -0.16  &  \em +11.18\%  &   \em +11.60\%\\
		\hline
		\multirow{3}{*}{\rotatebox[origin=c]{90}{\bf En-Zh}}
		& Standard   & 32.24 & 32.49K & 38.55 \\
        & Simplified & 33.15 & 36.59K & 54.06 \\
        \cdashline{2-5}
        & $\triangle$ & \em +0.91  & \em  +12.62\%  &  \em  +40.23\%\\
		\hline
		\multirow{3}{*}{\rotatebox[origin=c]{90}{\bf En-Fr}}
		& Standard    & 40.39 & 68.28K & 58.97 \\
        & Simplified  & 40.07 & 76.03K & 67.23 \\
        \cdashline{2-5}
        & $\triangle$ & \em  -0.32  & \em  +11.35\%  & \em  +14.01\%\\
	\end{tabular}
	\caption{Performance of the simplified Base decoder. ``\#Train'' denotes the training speed (words per second) and ``\#Infer.'' denotes the inference speed (sentences per second). Results are averages of three runs.}
	\label{tab:simplified}
\end{table}

\paragraph{Simplified Decoder Results}
\footnotetext{As a comparison, the total number of parameters in Base and Big models are 62.9M and 213.9M respectively on En-De.}

\begin{table}[t]
	\centering
	\begin{tabular}{c||cc}
		{\bf Model} & {\bf Fluency} & {\bf Adequacy} \\
		\hline
		Standard (Base)   &  4.00   &   3.86   \\
		\hline
		Simplified (Base) &  4.01   &  3.87 \\
	\end{tabular}
	\caption{Human evaluation of translation performance of both standard and simplified decoders on 100 samples from En-Zh test set, on the scale of 1 to 5.}
	\label{tab:human_evaluate}
\end{table}

Table~\ref{tab:simplified} reports the translation performance of both architectures on all three major datasets, while Figure~\ref{fig:FFN_prob} illustrates the information evolution of both on WMT En-De.
We find the simplified model reaches comparable performance with only a minimal drop of 0.1-0.3 BLEU on En-De and En-Fr, while observing 0.9 BLEU gains on En-Zh.\footnote{Simplified models are trained with the same hyper-parameters as standard ones, which may be suboptimal as the number of parameters is significantly reduced.}
To further assess the translation performance, we manually evaluate 100 translations sampled from the En-Zh test set. On the scale of 1 to 5, we find that the simplified decoder obtains a fluency score of 4.01 and an adequacy score of 3.87, which is approximately equivalent to that of the standard decoder, i.e. 4.00 for fluency and 3.86 for adequacy (in Table~\ref{tab:human_evaluate}).

On the other hand, since the simplified decoder drops the operations (FeedForward) with most parameters (shown in Table~\ref{tab:parameters}), we also expect a significant increase on training and inference speeds.
From Table~\ref{tab:simplified}, we confirm a consistent boost of both training and inference speeds by approximately 11-14\%.
To demonstrate the robustness, we also confirm our findings under Transformer big settings~\cite{Vaswani:2017:NIPS}, whose results are shown in  Section~\ref{sec:app_big}.
The lower PPL in Figure~\ref{fig:FFN_prob} suggests that
the simplified model also contains consistently more source and target information across its stacking layers.

Our results demonstrate that a single Add\&Norm is indeed sufficient for IFM, and the simplified model reaches comparable performance with a significant parameter reduction and a noticable 11-14\% boost on training and inference speed.

\section{Related Work}
\paragraph{Interpreting Encoder Representations}
Previous studies generally focus on interpreting the encoder representations by evaluating how informative they are for various linguistic tasks~\cite{conneau2018acl,tenney2019you}, for both RNN models~\cite{shi2016does,belinkov2017neural,bisazza2018lazy,blevins2018deep} and Transformer models~\cite{raganato2018analysis,tang:2019:emnlp,tenney2019bert,yang2019assessing}.
Although they found that a certain amount of linguistic information is captured by encoder representations, it is still unclear how much encoded information is used by the decoder.
Our work bridges this gap by interpreting how the Transformer decoder exploits the encoded information.

\paragraph{Interpreting Encoder Self-Attention}
In recent years, there has been a growing interest in interpreting the behaviors of attention modules.
Previous studies generally focus on the self-attention in the encoder, which is implemented as multi-head attention.
For example,~\newcite{li2018multi} showed that different attention heads in the encoder-side self-attention generally attend to the same position.
\newcite{voita2019analyzing} and \newcite{Michel:2019:NeurIPS} found that only a few attention heads play consistent and often linguistically-interpretable roles, and others can be pruned.
\newcite{Geng:2020:ACL} empirically validated that a selective mechanism can mitigate the problem of word order encoding and structure modeling of encoder-side self-attention.
In this work, we investigated the functionalities of decoder-side  attention modules for exploiting  both source and target information.

\paragraph{Interpreting Encoder Attention}
The encoder-attention weights are generally employed to interpret the output predictions of NMT models.
Recently,~\newcite{Jain2019AttentionIN} showed that attention weights are weakly correlated with the contribution of source words to the prediction.
~\newcite{He:2019:EMNLP} used the integrated gradients to better estimate the contribution of source words.
Related to our work,~\newcite{Li:2019:ACL} and~\newcite{Tang:2019:RANLP} also conducted word alignment analysis on the same De-En and Zh-En datasets with Transformer models\footnote{We find our results are more similar to that of~\newcite{Tang:2019:RANLP}. Also, our results are reported on the En$\Rightarrow$De and En$\Rightarrow$Zh directions, while they report results in the inverse directions.}.
We use similar techniques to examine word alignment in our context; however, we also introduce a forced-decoding-based probing task to closely examine the information flow. 

\paragraph{Understanding and Improving NMT}

Recent work started to improve NMT based on the findings of interpretation.
For instance,
\newcite{belinkov2017neural,belinkov2018evaluating} pointed out that different layers prioritize different linguistic types, based on which~\newcite{Dou:2018:EMNLP} and~\newcite{Yang:2019:AAAI} simultaneously exposed all of these signals to the subsequent process.
\citet{dalvi2017understanding} explained why the decoder learns considerably less morphology than the encoder, and then explored to explicitly inject morphology in the decoder.
\citet{emelin2019widening} argued that the need to represent and propagate lexical features in each layer limits the model's capacity, and introduced gated shortcut connections between the embedding layer and each subsequent layer.
\newcite{Wang:2020:ACL} revealed that miscalibration remains a severe challenge for NMT during inference, and proposed a graduated label smoothing that can improve the inference calibration.
In this work, based on our information probing analysis, we simplified the decoder by removing the residual feedforward module in totality, with minimal loss of translation quality and a significant boost of both training and inference speeds.

\section{Conclusions}

In this paper, we interpreted NMT Transformer decoder by assessing the evolution of both source and target information across layers and modules.
To this end, we investigated the information functionalities of decoder components in the translation process. Experimental results on three major datasets revealed several findings that help understand the behaviors of Transformer decoder from different perspectives.
We hope that our analysis and findings could inspire architectural changes for further improvements, such as 1) improving the word alignment of higher SEMs by incorporating external alignment signals; 2) exploring the stacking order of SEM, TEM and IFM sub-layers, which may provide a more effective way to transform information; 3) further pruning redundant sub-layers for efficiency.

Since our analysis approaches are not limited to the Transformer model, it is also interesting to explore other architectures such as RNMT~\cite{chen2018best}, ConvS2S~\cite{gehring2017convolutional}, or on document-level NMT~\cite{D17-1300,wang2019one}.
In addition, our analysis methods can be applied to other sequence-to-sequence tasks such as summarization and grammar error correction, whose source and target sides are in the same language. We leave those tasks for future work.

\section{Acknowledgments}
Tadepalli acknowledges the support of
DARPA under grant number  N66001-17-2-4030. The authors thank the anonymous reviewers for their insightful and helpful comments.

\bibliography{emnlp2020}
\bibliographystyle{acl_natbib}

\clearpage

\appendix

\section{Additional Results}

\subsection{Implementation Details}
\label{sec:app_implement}
All transformer models are selected based on their loss on validation set, while evaluated and reported on the test set. For En-De and En-Fr models, we used newstest2013 as validation set and newstest2014 as test set.
For En-Zh models, we used newsdev2016 as validation set and newstest2017 as test set.

All three datasets follow the prepossessing steps from FairSeq\footnote{\url{ https://github.com/pytorch/fairseq/blob/master/examples/translation/prepare-wmt14en2de.sh}}, which uses Moses tokenizer\footnote{\url{https://github.com/moses-smt/mosesdecoder/blob/master/scripts/tokenizer/mosestokenizer/tokenizer.py}}, with a joint BPE of 40000 steps, while does not include lower-casing nor true-casing.

All models are evaluated with a beam size of 10.
Before evaluating the BLEU score, we apply a postprocessing step, where En-De and En-Fr generations apply compound word splitting\footnote{\url{https://gist.github.com/myleott/da0ea3ce8ee7582b034b9711698d5c16}}, and En-Zh generations apply Chinese word splitting (into Chinese characters).
All generations are then evaluated with Moses {\em multi-bleu.perl} script\footnote{\url{https://github.com/moses-smt/mosesdecoder/blob/master/scripts/generic/multi-bleu.perl}} against the golden references.

\subsection{Transformer Big Results}
\label{sec:app_big}
We also compare the performance of the standard and simplified decoder under Transformer Big setting.
Big models are trained on 4 NVIDIA V100 chips, where each is allocated with a batch size of 8,192 tokens.
Other training schedules and hyper-parameters are the same as standard~\cite{Vaswani:2017:NIPS}.
Also, our Transformer Base models are all trained with full precision (FP32), while Big models are all trained with half precision (FP16) for faster training.

Transformer Big results are shown in Table.~\ref{tab:big_simplified}. We could observe a more severe BLEU score drop with a more significant speed boosting under Big setting. This is very intuitive, compared to Base setting, the simplified decoder drops more parameters, while still trained under the same schedule as standard, thus escalating the training discrepancy. Unfortunately due to the resource limitation, we could not afford hyper-parameter tuning for Transformer.

\begin{table}
	\centering
	\begin{tabular}{c|c|rrr}
		\multicolumn{2}{c|}{\bf Decoder} & \bf BLEU & \bf \#Train. & \bf \#Infer. \\
		\hline
		\multirow{3}{*}{\rotatebox[origin=c]{90}{\bf En-De}}
		& Standard   & 28.66 & 103.7K &  74.3\\
        & Simplified & 28.20 & 125.2K &  90.5\\
        \cdashline{2-5}
        & $\triangle$ & \em -0.46  &  \em +20.7\%  &   \em +21.8\%\\
		\hline
		\multirow{3}{*}{\rotatebox[origin=c]{90}{\bf En-Zh}}
		& Standard   & 34.48 & 71.3K &  30.5\\
        & Simplified & 34.35 & 82.6K &  46.0\\
        \cdashline{2-5}
        & $\triangle$ & \em -0.13  & \em  +15.8\%  &  \em  +50.8\%\\
		\hline
		\multirow{3}{*}{\rotatebox[origin=c]{90}{\bf En-Fr}}
		& Standard    & 42.48 & 113.8K &  65.7\\
        & Simplified  & 42.19 & 138.1K &  80.9\\
        \cdashline{2-5}
        & $\triangle$ & \em  -0.29  & \em  +21.4\%  & \em  +23.1\%\\
	\end{tabular}
	\caption{Performance of the simplified Big decoder. ``\#Train'' denotes the training speed (words per second) and ``\#Infer.'' denotes the inference speed (sentences per second).}
	\label{tab:big_simplified}
\end{table}

\subsection{Additional En-Zh and En-Fr Plots}
\label{sec:app_enzh_enfr}
All experiments are conducted on three datasets (En-De, En-Zh and En-Fr), where we have similar findings. Due to space limits, we mainly demonstrate results on En-De task in our paper. In this section, we provide additional results on En-Zh and En-Fr if applicable.

\begin{figure}[t]
	\centering
	\subfloat[Source PPL]{
		\includegraphics[height=0.3\textwidth]{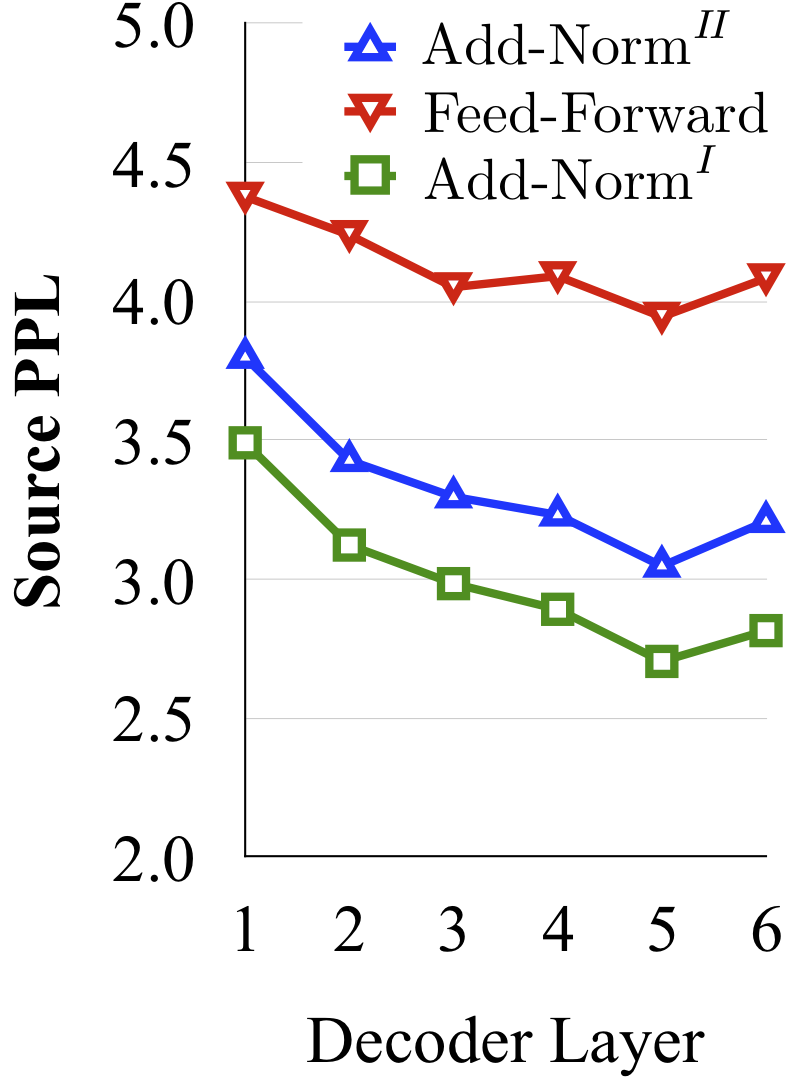}}
	\hfill
	\subfloat[Target PPL]{
		\includegraphics[height=0.3\textwidth]{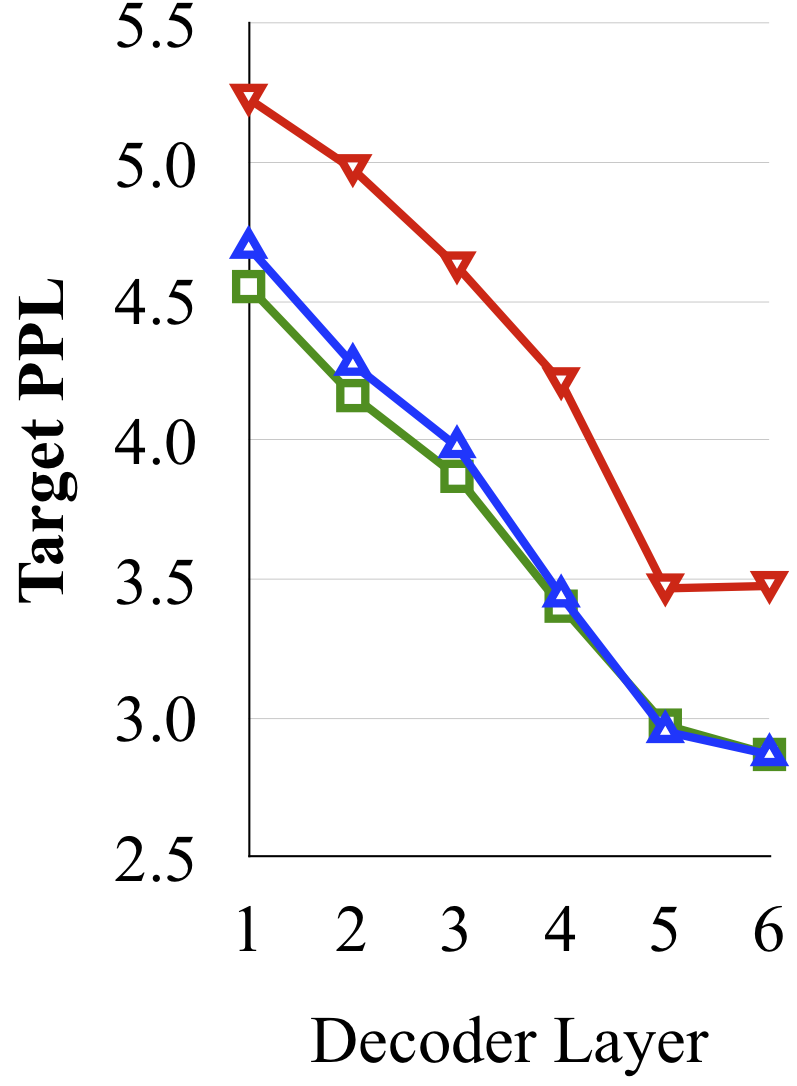}}
	\caption{Illustration of the source and target information evolution within IFM on En-Zh.}
	\label{fig:IFM_prob_enzh}
\end{figure}

\begin{figure}[t]
	\centering
	\subfloat[Word Alignment]{
		\includegraphics[height=0.3\textwidth]{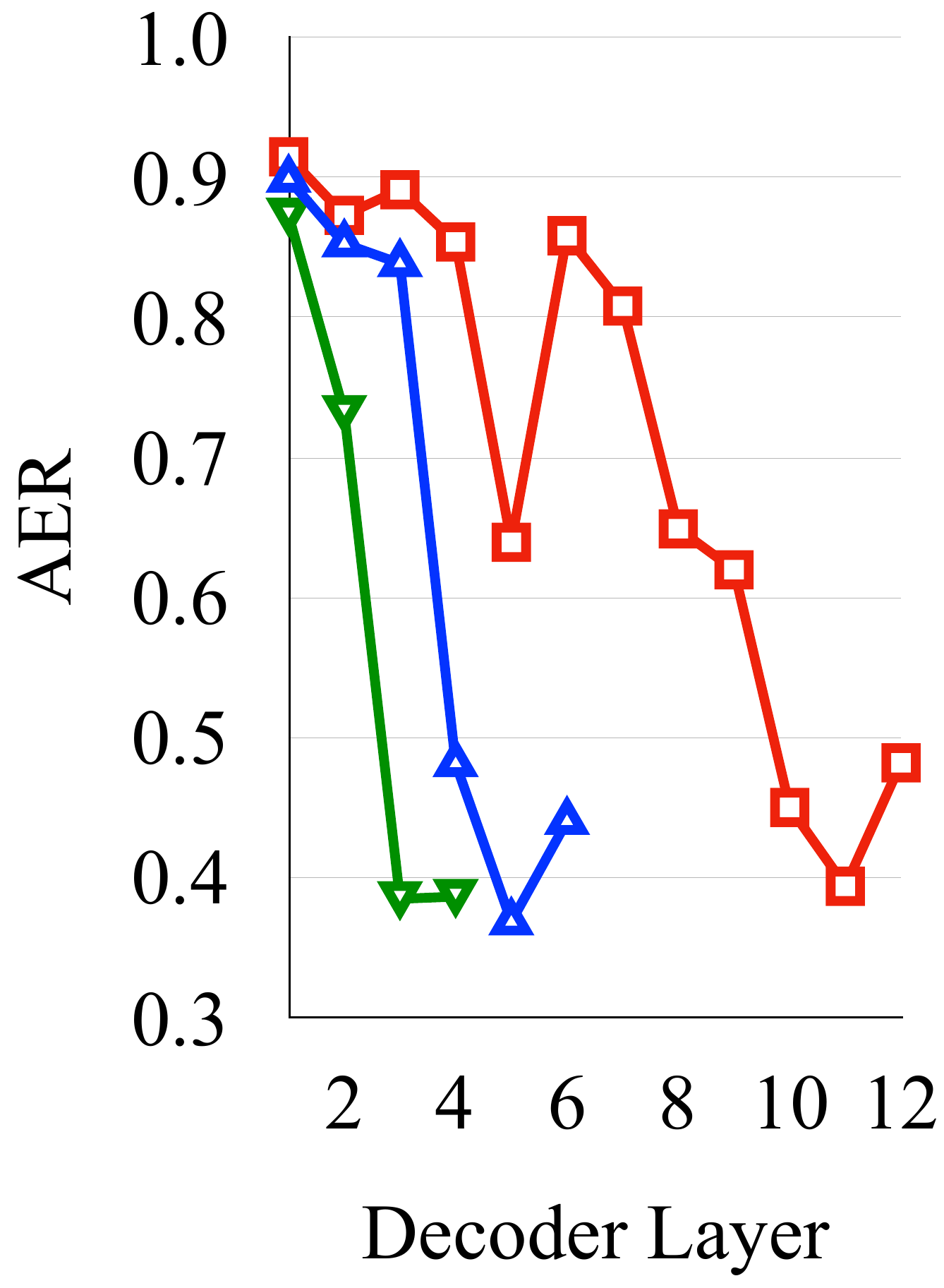}}
	\hfill
	\subfloat[Cumulative Coverage]{
		\includegraphics[height=0.3\textwidth]{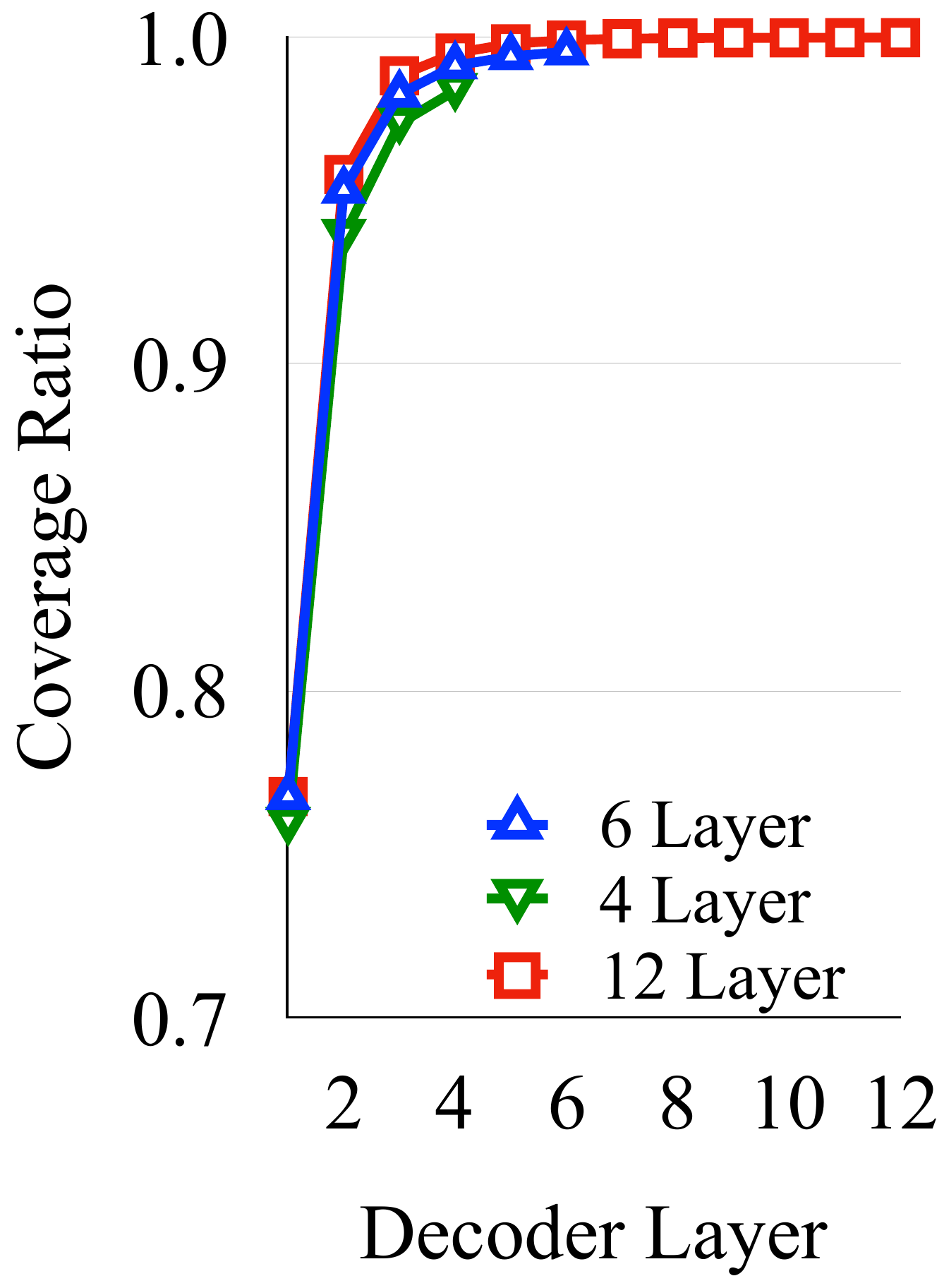}}
	\caption{Effects of decoder depths on SEM behaviors on En-Zh.}
	\label{fig:depth_enzh}
\end{figure}

\begin{figure}[t]
	\centering
	\subfloat[Word Alignment]{
		\includegraphics[height=0.3\textwidth]{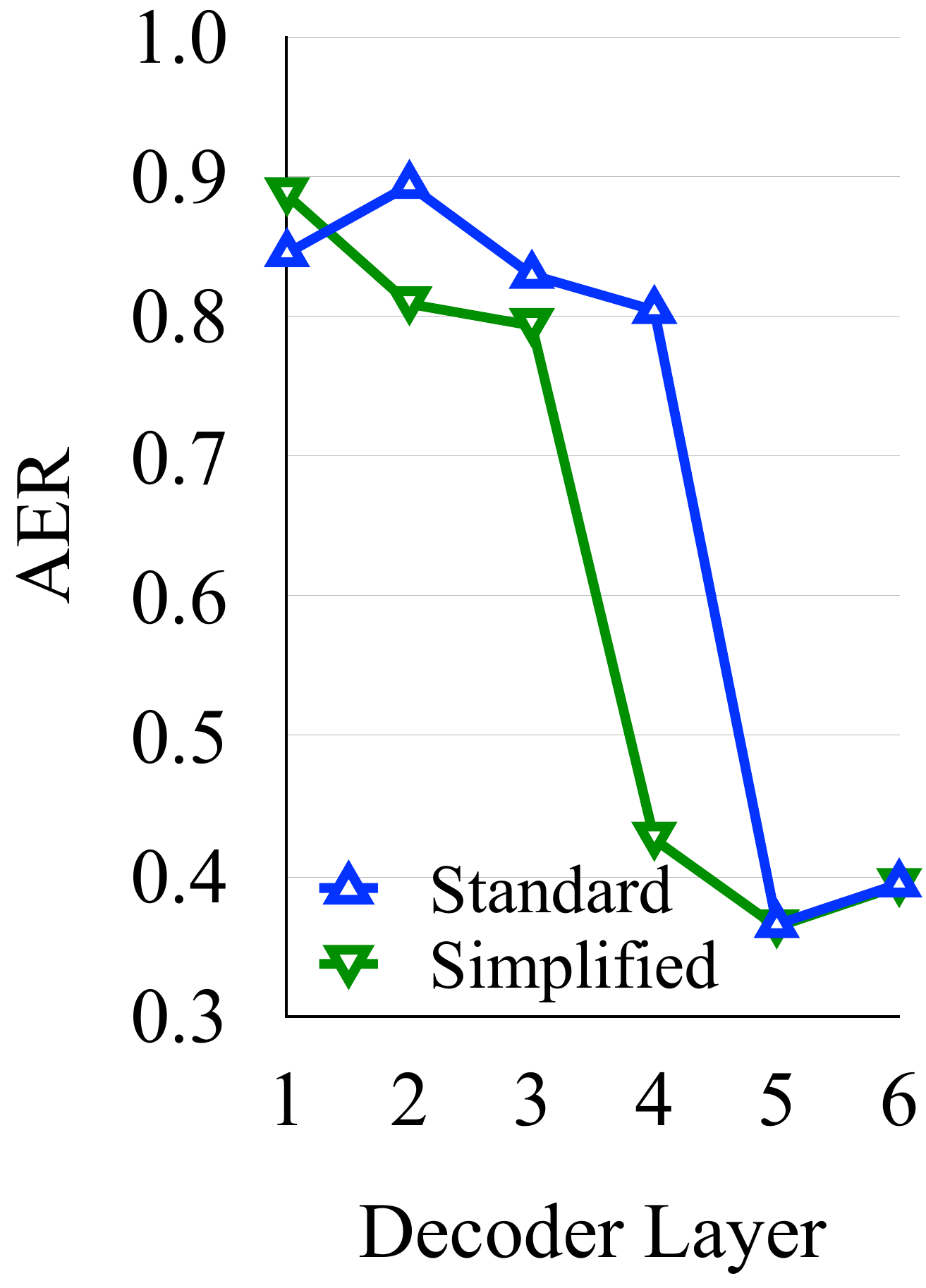}}
	\hfill
	\subfloat[Cumulative Coverage]{
        \includegraphics[height=0.3\textwidth]{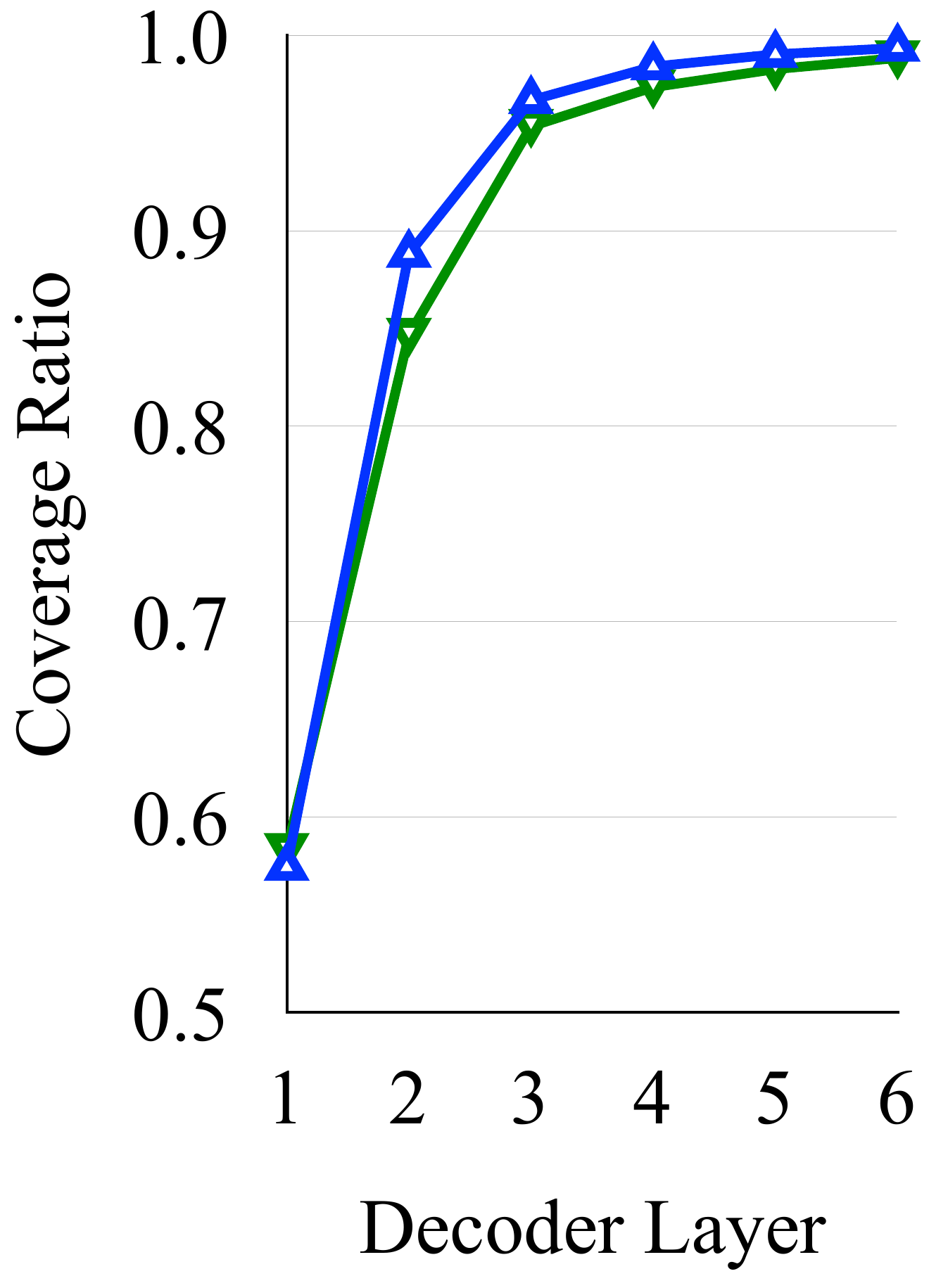}}
	\caption{Comparison between standard and simplified model on SEM behaviors on En-De.}
	\label{fig:simplify_aer_ende}
\end{figure}

\begin{figure}[t]
	\centering
	\subfloat[Word Alignment]{
		\includegraphics[height=0.3\textwidth]{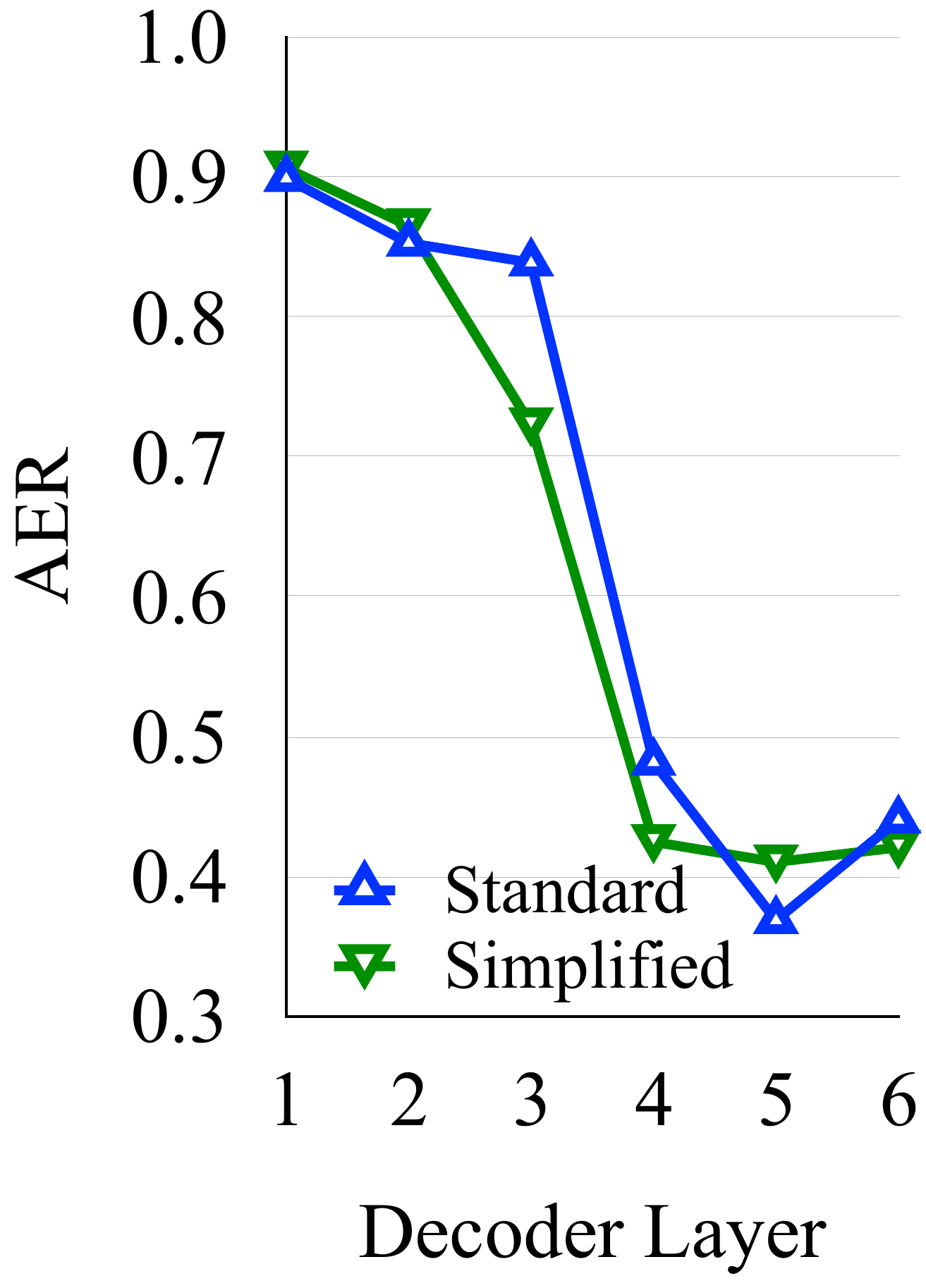}}
	\hfill
	\subfloat[Cumulative Coverage]{
		\includegraphics[height=0.3\textwidth]{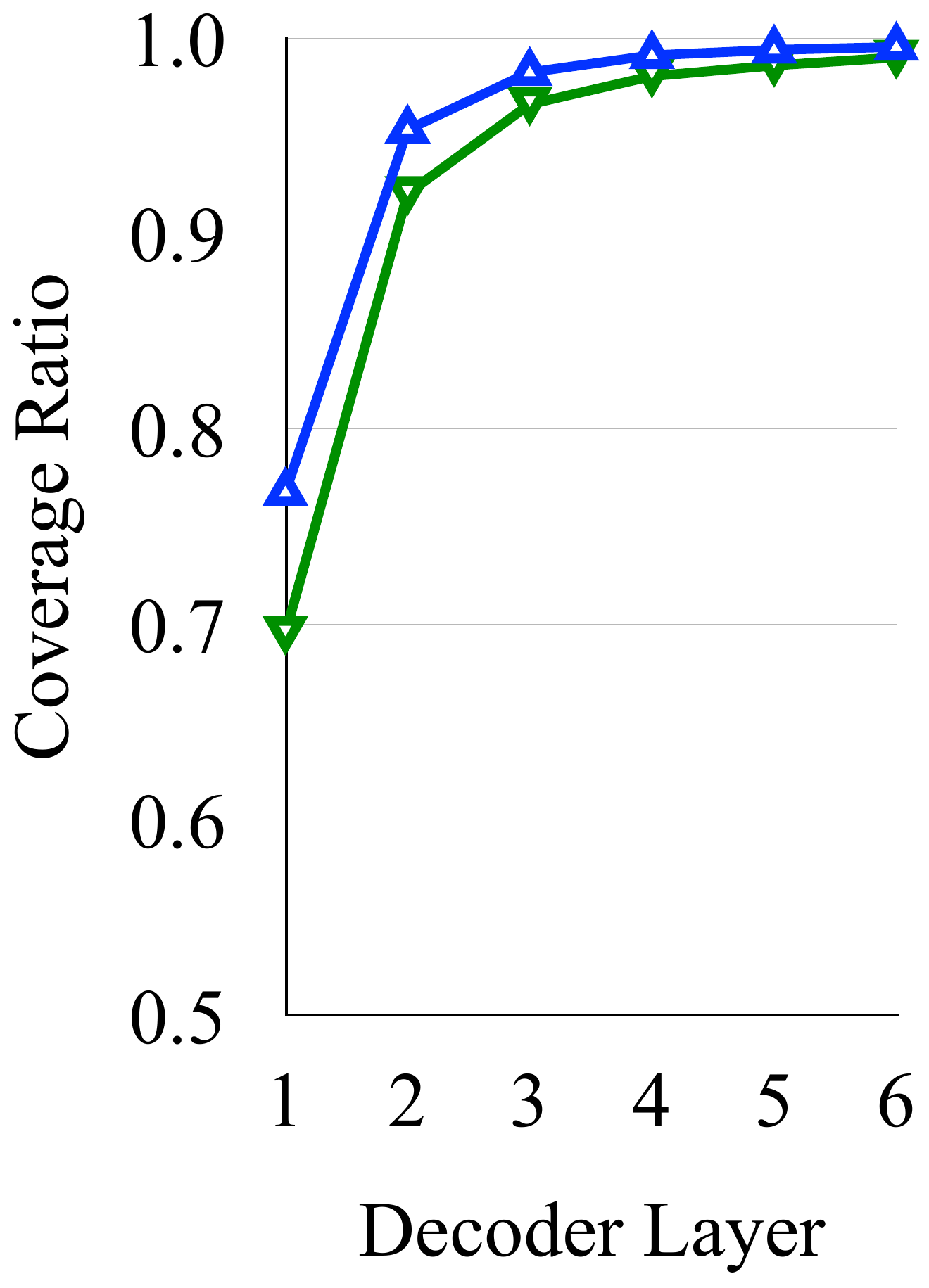}}
	\caption{Comparison between standard and simplified model on SEM behaviors on En-Zh.}
	\label{fig:simplify_aer_enzh}
\end{figure}

\end{document}